\pdfoutput=1

\documentclass[11pt]{article}

\usepackage[preprint]{acl}

\usepackage{times}
\usepackage{latexsym}

\usepackage[T1]{fontenc}
\usepackage{CJKutf8}

\usepackage[utf8]{inputenc}

\usepackage{microtype}

\usepackage{inconsolata}

\usepackage{graphicx}

\usepackage{inconsolata}

\usepackage{hyperref}
\usepackage{transparent}
\usepackage{array}
\usepackage{multirow}
\usepackage{multicol}
\usepackage{xprintlen}
\usepackage{csquotes}
\usepackage[inline]{enumitem}
\usepackage[normalem]{ulem}
\usepackage[export]{adjustbox}
\usepackage{float}
\usepackage{placeins}
\usepackage{amsmath}
\usepackage{makecell}
\usepackage{colortbl}
\usepackage{siunitx}
\usepackage{ragged2e}
\usepackage{tcolorbox}
\usepackage{booktabs}
\usepackage{bbding}
\usepackage{soul}

\usepackage{subcaption}  %

\usepackage{letltxmacro}

\AtBeginDocument{%
  \LetLtxMacro\origcite\cite
  \LetLtxMacro\origcitep\citep
  \LetLtxMacro\origcitet\citet
}

\newcolumntype{P}[1]{>{\centering\arraybackslash}m{#1}}

\newcommand{\ndnote}[1]{{\color{purple}ND:[#1]}}

\definecolor{darkgreen}{HTML}{14781c}

\renewcommand{\ndnote}[1]{}

\def\gl#1{~\textcolor{gray}{\small(#1)}}

\NewDocumentCommand\cancutwrap{m o o m}{\mbox{%
  \IfNoValueTF{#2}{#1{#4}}{%
    \IfNoValueTF{#3}{#1[#2]{#4}}{#1[#2][#3]{#4}}}}}

\newcommand{\cancut}[1]{\textcolor{gray}{%
  \def\cite{\cancutwrap\origcite}%
  \def\citep{\cancutwrap\origcitep}%
  \def\citet{\cancutwrap\origcitet}%
  \sout{#1}}}

\renewcommand{\cancut}[1]{}   

\definecolor{babyblue}{rgb}{0.54, 0.81, 0.94}
\definecolor{peach}{rgb}{.9, .6, .36}
\definecolor{lightgreen}{rgb}{.37, .63, .47}
\newcommand{\hlcolor}[3]{{\sethlcolor{#2!#3}\hl{#1}}}

\newcommand{\mosaic}{\textsc{Mosaic-Emo}}

\definecolor{nacell}{rgb}{0.57, 0.57, 0.57}

\title{Examining Multilingual Emotion Network Topology and Probing Its Underlying Valence-Arousal Structure}

\author{
 \textbf{Jennifer Jane Ries},
 \textbf{Nicholas Deas},
 \textbf{Kathleen McKeown}
\\
\\
 Department of Computer Science, Columbia University \\
\\
 \small{
   \href{mailto:iep2110@columbia.edu}{jr3954@columbia.edu},
   \href{mailto:ndeas@cs.columbia.edu}{[ndeas,kathy]@cs.columbia.edu} 
 }
}

\begin{document}
\maketitle

\begin{abstract}

As large language models (LLMs) interact with an increasingly wide and diverse group of users, it is critical that LLMs represent emotion concepts in culturally appropriate ways. We conduct an exploratory analysis on monolingual and multilingual networks based on similarity between token-level representations of affective state terms in three LLMs: Qwen3-8B, Ministral-3-8B-Base-2512 and Gemma-3-12B-pt.
Using Louvain community detection on monolingual networks, we first surface language-specific differences in emotion space structure, and determine that the semantic relationships encoded between emotion concepts largely match human conceptual similarities reflected in prior linguistic work. In cross-lingual networks, we find that
networks show cohesive, conceptually related multilingual emotion communities. In order to better understand our network results, we then examine to what extent these models encode language-specific valence and arousal structure in models' hidden representations using Minimum Description Length probes. Thus, we present a systematic analysis of fine-grained emotion representation structure across languages through both inductive and deductive lenses.

\end{abstract}

\definecolor{C0}{HTML}{4477AA}
\definecolor{C1}{HTML}{EE6677}
\definecolor{C2}{HTML}{228833}
\definecolor{C3}{HTML}{CCBB44}
\definecolor{C4}{HTML}{66CCEE}
\definecolor{C5}{HTML}{AA3377}
\definecolor{C6}{HTML}{BBBBBB}

\newcommand{\dataStats}{
    \begin{table}[htbp]
        \centering
        \resizebox{0.95\linewidth}{!}{
            \begin{tabular}{cc ccc cccc}
                \toprule
                \multirow{2}{*}{\textbf{Lang.}} & \multirow{2}{*}{\textbf{Split}} & \multirow{2}{*}{\textbf{Size}} & \multirow{2}{*}{\shortstack{\textbf{Unq.}\\\textbf{States}}} & \multirow{2}{*}{\textbf{Len.}}  & \multicolumn{4}{c}{\textbf{Score Dist. (\%)}} \\
                \cmidrule(lr){6-9}
                                            &          &  & &   & 0 & 1 & 2 & 3 \\
                \midrule
                \multirow{3}{*}{En}         & Exs & 150 & 106 & 22.8 & 8.0 & 2.7 & 5.3 & 84.0 \\
                                            & Test & 49 & 41 & 26.4 & 8.2 & 6.1 & 26.5 & 59.2 \\
                \cmidrule(lr){2-9}
                                            & Ovr & 199 & 131 & 23.7 & 8.0 & 3.5 & 10.6 & 77.9 \\
                 \midrule
                \multirow{3}{*}{Es}         & Exs & 150 & 113 & 63.8 & 27.3 & 3.3 & 8.0 & 61.3 \\
                                            & Test & 50 & 45 & 68.5 & 20.0 & 2.0 & 22.0 & 56.0 \\
                \cmidrule(lr){2-9}
                                            & Ovr & 200 & 143 & 65.0 & 25.5 & 3.0 & 11.5 & 60.0 \\
                \bottomrule
            \end{tabular}
        }
        \caption{Summary data statistics with (rounded) percentage of words annotated within each score range by human annotators. \textit{Exs} represents the single-coded samples used as few-shot examples, while \textit{Test} represents the double-coded samples used for all evaluations. \textit{Unq. States} represents count of unique affective states.~\protect\footnotemark}
        \label{tab:data-stats}
    \end{table}
}

\newcommand{\dataExamples}{

    \begin{table}[htbp]
        \centering
        \resizebox{\linewidth}{!}{
            \begin{tabular}{c P{8cm} c }
                \toprule
                  Lang & Context & Score \\
                 \midrule
                 \multirow{2}{*}{En}            & I feel \hlcolor{jittery}{C1}{80} waking up but its probably left over anxiety. & 1 \\
                 \cmidrule(lr){2-3}
                                                & I’m so anxious, hoping I’m paranoid, but pissed of that I feel \hlcolor{manipulated}{C1}{80} into being fearful vs. having a normal relationship with someone. & 3   \\
                 \midrule
                 \multirow{2}{*}{Es}            & Tengo una hipoteca. Odio a los corruptos, la falta de meritocracia y la injusticia. Estoy \hlcolor{rodeado}{C1}{80} de gente en paro, de empresas que no han sido pagadas por los servicios prestados y de hipotecados. & 0 \\
                 \cmidrule(lr){2-3}
                                                & He tratado de hacerlo muchas veces pero nunca ha dado resultado, incluso llendo a bailar con pololas o amigos el resultado es el mismo, nose donde poner mis brazos ni como moverme, me siento \hlcolor{incomodo}{C1}{80} & 3   \\
                \bottomrule
            \end{tabular}
        }
        \caption{Example texts and affect ratings on a 4pt-Likert scale, higher being "more like an affective state." }
        \label{tab:ex-annot}
    \end{table}

}

\newcommand{\verificationPromptRating}{

    \begin{figure}[htbp]
        \centering
        \begin{tcolorbox}[colback=gray!20, colframe=black, width=\columnwidth]
            \small
            \textbf{Affective State Verification Prompt:} \\[5pt]
            \hrule
            \vspace{5pt}
            \textbf{System:} You are an expert in human emotions and feelings. \\
            \hrule
            \vspace{5pt}
            \textbf{User:} Affective state refers to  any terms that humans use to describe their experiences of feeling, including emotions, moods, and figurative expressions of feelings (e.g. 'blue' as an expression of sadness instead of the color). Does the term between <span> and </span> reflect an affective state? Answer only with one character from the following: 0, 1, 2, or 3.
            
            0 means Not an affective state: the term does not refer to an emotion, feeling, or internal state.
            
            1 means Unlike an affective state: the term is referencing something that is not an emotion.
            
            2 means Like an affective state: the term is likely referencing an emotion, feeling, or internal state.
            
            3 means Completely an affective state: the term is definitely an emotion, feeling, or internal state.
            
            Do not explain or preface your answer.\\
            
            {\color{blue}%
            Input text:  Day 7 with Covid, I am feeling <span>much</span> better 
            
            Response: 0\\%
            }
            
            {\color{red}%
            Input text: I feel <span>miserable</span> and overwhelmed but I know that everyone will look down on me if I don't get a degree or if I drop out of school.
            
            Response:%
            }
        \end{tcolorbox}
        \caption{Full rating prompt template provided to models for affective state verification experiments.}
        \label{fig:full-ver-prompt-rating}
        \vspace{-1em}
    \end{figure}

}

\newcommand{\verificationPromptYN}{

    \begin{figure}[htbp]
        \centering
        \begin{tcolorbox}[colback=gray!20, colframe=black, width=\columnwidth]
            \small
            \textbf{Affective State Verification Prompt:} \\[5pt]
            \hrule
            \vspace{5pt}
            \textbf{System:} You are an expert in human emotions and feelings. \\
            \hrule
            \vspace{5pt}
            \textbf{User:} Affective state refers to  any terms that humans use to describe their experiences of feeling, including emotions, moods, and figurative expressions of feelings (e.g. 'blue' as an expression of sadness instead of the color). Tell me if the word between <span> and </span> is an affective state or not. Output only Y if it is an affective state or N if it is not.\\
            Do not explain or preface your answer.\\
            
            {\color{blue}%
            Input text:  Day 7 with Covid, I am feeling <span>much</span> better 
            
            Response: 0\\%
            }
            
            {\color{red}%
            Input text: I feel <span>miserable</span> and overwhelmed but I know that everyone will look down on me if I don't get a degree or if I drop out of school.
            
            Response:%
            }
        \end{tcolorbox}
        \caption{Full binary prompt template provided to models for affective state verification experiments.}
        \label{fig:full-ver-prompt-yn}
        \vspace{-1em}
    \end{figure}

}

\newcommand{\condensedVerificationPrompt}{

    \begin{figure}[htbp]
        \centering
        \begin{tcolorbox}[colback=gray!20, colframe=black, width=\columnwidth]
            \small
            \textbf{Affective State Verification Prompt:} \\[5pt]
            \hrule
            \vspace{5pt}
            Affective state refers to  any terms that humans use to describe their experiences of feeling, including emotions, moods, and figurative expressions of feelings ... Does the term between <span> and </span> reflect an affective state? Answer only with one character from the following: 0, 1, 2, or 3.
            
           ...
            
            \hlcolor{%
            Input text:  Day 7 with Covid, I am feeling <span>much</span> better 
            
            Response: 0\\%
            }{C0}{80}
            
            \hlcolor{%
            Input text: I feel <span>}{C1}{80}\hlcolor{miserable}{C2}{80}\hlcolor{</span> and overwhelmed but I know that everyone will look down on me if I don't get a degree or if I drop out of school.
            
            Response:%
            }{C1}{80}
        \end{tcolorbox}
        \caption{Truncated prompt template provided to models for affective state verification experiments. \hlcolor{Blue text}{C0}{80} represents in-context examples provided to models, \hlcolor{red text}{C1}{80} represents the test example in context, and \hlcolor{green text}{C2}{80} indicates the target affective state.}
        \label{fig:ver-prompt}
    \end{figure}

}

\newcommand{\condensedExpressionPrompt}{

    \begin{figure}[htbp]
        \centering
        \begin{tcolorbox}[colback=gray!20, colframe=black, width=\columnwidth]
            \small
            \textbf{Affective State Expression Prompt:} \\[5pt]
            \hrule
            \vspace{5pt}
            Write a sample response from someone that is feeling \hlcolor{miserable}{C2}{80} to the following fictional user message. The sample response should clearly express that the author feels \hlcolor{miserable}{C2}{80}. Do not, however, explicitly mention "\hlcolor{miserable}{C2}{80}" or use asterisks to indicate actions.
            
            User Prompt: \hlcolor{I'm feeling really down today and need a pick me up.}{C1}{80}
        \end{tcolorbox}
        \caption{Truncated prompt template provided to models for affective state expression experiments using the rating output format. \hlcolor{Red text}{C1}{80} represents the user prompt the model is asked to respond to, while the \hlcolor{green text}{C2}{80} represents the target affective state that should be expressed in the model's response.}
        \label{fig:exp-prompt}
    \end{figure}

}

\newcommand{\condensedIdentificationPrompt}{

    \begin{figure}[htbp]
        \centering
        \begin{tcolorbox}[colback=gray!20, colframe=black, width=\columnwidth]
            \small
            \textbf{Affective State Identification Prompt:} \\[5pt]
            \hrule
            \vspace{5pt}
            Determine the most likely term reflecting a feeling to replace the <MASK> token(s) in the following text: "{input_context}" Do not introduce the answer, respond ONLY with a single lowercase term.
            
            User Prompt: \hlcolor{I'm feeling really down today and need a pick me up.}{C1}{80}
        \end{tcolorbox}
        \caption{Truncated prompt template provided to models for affective state expression experiments using the rating output format. \hlcolor{Red text}{C1}{80} represents the user prompt the model is asked to respond to, while the \hlcolor{green text}{C2}{80} represents the target affective state that should be expressed in the model's response. }
        \label{fig:exp-prompt}
    \end{figure}

}

\newcommand{\allCondensedPrompts} {

    \begin{figure*}[htbp]
        \centering
        \begin{subfigure}[c]{0.245\textwidth}
            \begin{tcolorbox}[colback=gray!20, colframe=black, width=\columnwidth]
                \scriptsize
                Determine the most likely term reflecting a feeling to replace the <MASK> token(s) in the following text: "\hlcolor{I feel }{C1}{80}\hlcolor{<MASK>}{C2}{80}\hlcolor{</span> and overwhelmed but I know that everyone will look down on me if I don't get a degree or if I drop out of school.}{C1}{80}" \\

                Do not introduce the answer, respond ONLY with a single lowercase term.\\
            \end{tcolorbox}
            \vspace{-1em}
            \caption{\textbf{Identification}}
            \label{fig:ident-prompt}
        \end{subfigure}
        \hfill
        \begin{subfigure}[c]{0.25\textwidth}
            \begin{tcolorbox}[colback=gray!20, colframe=black, width=\columnwidth]
                \scriptsize
                Write a sample response from someone that is feeling \hlcolor{miserable}{C2}{80} to the following fictional user message. The sample response should clearly express that the author feels \hlcolor{miserable}{C2}{80}. Do not, however, explicitly mention "\hlcolor{miserable}{C2}{80}" or use asterisks to indicate actions.\\

                User Prompt: \hlcolor{I'm feeling really down today and need a pick me up.}{C1}{80}
            \end{tcolorbox}
            \vspace{-1em}
            \caption{\textbf{Expression}}
            \label{fig:exp-prompt}
        \end{subfigure}%
        \hfill
        \begin{subfigure}[c]{0.47\textwidth}
            \begin{tcolorbox}[colback=gray!20, colframe=black, width=\columnwidth]
                \scriptsize
                Affective state refers to  any terms that humans use to describe their experiences of feeling, including emotions, moods, and figurative expressions of feelings ... Does the term between <span> and </span> reflect an affective state? Answer only with one character from the following: 0, 1, 2, or 3.
                
               ...
                
                \hlcolor{%
                Input text:  Day 7 with Covid, I am feeling <span>much</span> better 
                
                Response: 0\\%
                }{C0}{80}
                
                \hlcolor{%
                Input text: I feel <span>}{C1}{80}\hlcolor{miserable}{C2}{80}\hlcolor{</span> and overwhelmed but I know that everyone will look down on me if I don't get a degree or if I drop out of school.
                
                Response:%
                }{C1}{80}
            \end{tcolorbox}
            \vspace{-1em}
            \caption{\textbf{Verification}}
            \label{fig:ver-prompt}
        \end{subfigure}%
        \caption{Condensed prompt templates for the three tasks. \hlcolor{Red text}{C1}{80} represents input texts drawn from MASIVE, \hlcolor{green text}{C2}{80} represents a masked or target affective state term, and \hlcolor{blue text}{C0}{80} represents in-context examples in the verification task. }
        \label{fig:task-prompts}
        \vspace{-1em}
    \end{figure*}

}

\newcommand{\condensedIdentificationResults}{
\begin{table}[htbp]
    \centering
    \resizebox{0.95\linewidth}{!}{
        \begin{tabular}{ccccc}
            \toprule
            & \multirow{2}{*}{Model} & \multicolumn{3}{c}{Top-$k$ Similarity $\uparrow$} \\
            \cmidrule(lr){3-5}
            & & 1 & 3 & 5 \\
            \midrule
            \cellcolor{gray!20} & OLMo 3 Base & \cellcolor{C5!35}$37.1\% \pm \text{\small11.3}$ & \cellcolor{C5!24}$37.3\% \pm \text{\small11.4}$ & \cellcolor{C5!22}$37.3\% \pm \text{\small11.4}$ \\
            \cellcolor{gray!20} & OLMo 3 & \cellcolor{C5!70}$46.2\% \pm \text{\small19.4}$ & \cellcolor{C5!70}$58.6\% \pm \text{\small20.2}$ & \cellcolor{C5!70}$63.0\% \pm \text{\small19.9}$ \\
            \cellcolor{gray!20} & Llama 3.1 Base & \cellcolor{C5!10}$30.3\% \pm \text{\small9.7}$ & \cellcolor{C5!10}$30.4\% \pm \text{\small9.6}$ & \cellcolor{C5!10}$30.4\% \pm \text{\small9.6}$ \\
            \cellcolor{gray!20} & Llama 3.1 & \cellcolor{C5!64}$44.8\% \pm \text{\small13.6}$ & \cellcolor{C5!54}$51.5\% \pm \text{\small15.2}$ & \cellcolor{C5!57}$56.0\% \pm \text{\small16.9}$ \\
            \cellcolor{gray!20} & Qwen 3 Base & \cellcolor{C5!24}$34.2\% \pm \text{\small14.0}$ & \cellcolor{C5!26}$38.1\% \pm \text{\small17.2}$ & \cellcolor{C5!24}$38.3\% \pm \text{\small17.1}$ \\
            \cellcolor{gray!20} & Qwen 3 & \cellcolor{C5!68}$45.8\% \pm \text{\small13.9}$ & \cellcolor{C5!64}$55.8\% \pm \text{\small16.0}$ & \cellcolor{C5!60}$57.7\% \pm \text{\small16.4}$ \\
            \cellcolor{gray!20} & Ministral 3 Base & \cellcolor{C5!10}$30.3\% \pm \text{\small8.4}$ & \cellcolor{C5!10}$30.8\% \pm \text{\small8.5}$ & \cellcolor{C5!10}$30.9\% \pm \text{\small8.5}$ \\
            \multirow{-8}{*}{\cellcolor{gray!20}\makecell{\rotatebox{90}{English}}} & Ministral 3 & \cellcolor{C5!58}$43.3\% \pm \text{\small15.2}$ & \cellcolor{C5!59}$53.7\% \pm \text{\small16.9}$ & \cellcolor{C5!61}$58.4\% \pm \text{\small16.6}$ \\
            \midrule
            \cellcolor{gray!20} & OLMo 3 Base & \cellcolor{C5!10}$30.7\% \pm \text{\small8.0}$ & \cellcolor{C5!10}$30.9\% \pm \text{\small8.1}$ & \cellcolor{C5!10}$30.9\% \pm \text{\small8.1}$ \\
            \cellcolor{gray!20} & OLMo 3 & \cellcolor{C5!58}$42.9\% \pm \text{\small14.2}$ & \cellcolor{C5!50}$49.4\% \pm \text{\small13.4}$ & \cellcolor{C5!54}$53.3\% \pm \text{\small14.4}$ \\
            \cellcolor{gray!20} & Llama 3.1 Base & \cellcolor{C5!46}$39.8\% \pm \text{\small10.6}$ & \cellcolor{C5!29}$40.0\% \pm \text{\small10.7}$ & \cellcolor{C5!28}$40.1\% \pm \text{\small10.6}$ \\
            \cellcolor{gray!20} & Llama 3.1 & \cellcolor{C5!64}$44.3\% \pm \text{\small16.8}$ & \cellcolor{C5!57}$52.5\% \pm \text{\small17.2}$ & \cellcolor{C5!61}$56.8\% \pm \text{\small18.3}$ \\
            \cellcolor{gray!20} & Qwen 3 Base & \cellcolor{C5!37}$37.7\% \pm \text{\small10.9}$ & \cellcolor{C5!27}$38.9\% \pm \text{\small11.0}$ & \cellcolor{C5!28}$40.2\% \pm \text{\small11.3}$ \\
            \cellcolor{gray!20} & Qwen 3 & \cellcolor{C5!49}$40.6\% \pm \text{\small15.5}$ & \cellcolor{C5!50}$49.5\% \pm \text{\small15.1}$ & \cellcolor{C5!60}$56.2\% \pm \text{\small17.4}$ \\
            \cellcolor{gray!20} & Ministral 3 Base & \cellcolor{C5!40}$38.3\% \pm \text{\small10.8}$ & \cellcolor{C5!26}$38.4\% \pm \text{\small10.8}$ & \cellcolor{C5!24}$38.4\% \pm \text{\small10.8}$ \\
            \multirow{-8}{*}{\cellcolor{gray!20} \makecell{\rotatebox{90}{Spanish}}} & Ministral 3 & \cellcolor{C5!70}$45.7\% \pm \text{\small19.8}$ & \cellcolor{C5!70}$58.4\% \pm \text{\small20.3}$ & \cellcolor{C5!70}$60.9\% \pm \text{\small21.1}$ \\
            \bottomrule
        \end{tabular}
    }
    \caption{Top-k similarity scores with $\pm$ 2SE's for a subset of models in the identification task. 
    Scores are shaded relative to those of other models for each value of $k$ (darker indicating better performance).
    }
    \label{tab:ident-benchmark}
\end{table}
}

\newcommand{\fullIdentificationResults}{
\begin{table}[htbp]
    \centering
    \resizebox{0.95\linewidth}{!}{
        \begin{tabular}{ccccc}
            \toprule
            & \multirow{2}{*}{Model} & \multicolumn{3}{c}{Top-$k$ Similarity$\uparrow$} \\
            \cmidrule(lr){3-5}
            & & 1 & 3 & 5 \\
            \midrule
            \cellcolor{gray!20} & Llama 2 & \cellcolor{C5!40}$37.0\% \pm \text{\small13.8}$ & \cellcolor{C5!35}$40.4\% \pm \text{\small14.5}$ & \cellcolor{C5!31}$40.8\% \pm \text{\small14.6}$ \\
            \cellcolor{gray!20} & EmoLlama 2 & \cellcolor{C5!45}$39.2\% \pm \text{\small12.8}$ & \cellcolor{C5!44}$45.1\% \pm \text{\small16.5}$ & \cellcolor{C5!39}$45.4\% \pm \text{\small16.4}$ \\
            \cellcolor{gray!20} & OLMo & \cellcolor{C5!26}$31.1\% \pm \text{\small10.7}$ & \cellcolor{C5!23}$33.9\% \pm \text{\small10.5}$ & \cellcolor{C5!22}$35.9\% \pm \text{\small11.8}$ \\
            \cellcolor{gray!20} & OLMo 2 & \cellcolor{C5!58}$44.2\% \pm \text{\small14.4}$ & \cellcolor{C5!57}$52.0\% \pm \text{\small14.7}$ & \cellcolor{C5!58}$56.4\% \pm \text{\small13.7}$ \\
            \cellcolor{gray!20} & OLMo 3 Base & \cellcolor{C5!40}$37.1\% \pm \text{\small11.3}$ & \cellcolor{C5!29}$37.3\% \pm \text{\small11.4}$ & \cellcolor{C5!24}$37.3\% \pm \text{\small11.4}$ \\
            \cellcolor{gray!20} & OLMo 3 & \cellcolor{C5!63}$46.2\% \pm \text{\small19.4}$ & \cellcolor{C5!70}$58.6\% \pm \text{\small20.2}$ & \cellcolor{C5!70}$63.0\% \pm \text{\small19.9}$ \\
            \cellcolor{gray!20} & Deepseek (Llama) & \cellcolor{C5!11}$25.2\% \pm \text{\small12.1}$ & \cellcolor{C5!14}$29.2\% \pm \text{\small11.2}$ & \cellcolor{C5!12}$30.2\% \pm \text{\small11.0}$ \\
            \cellcolor{gray!20} & Deepseek (Qwen) & \cellcolor{C5!10}$24.5\% \pm \text{\small6.3}$ & \cellcolor{C5!10}$26.8\% \pm \text{\small6.9}$ & \cellcolor{C5!10}$28.7\% \pm \text{\small7.6}$ \\
            \cellcolor{gray!20} & Llama 3 Base & \cellcolor{C5!27}$31.9\% \pm \text{\small9.2}$ & \cellcolor{C5!19}$32.1\% \pm \text{\small9.4}$ & \cellcolor{C5!15}$32.1\% \pm \text{\small9.4}$ \\
            \cellcolor{gray!20} & Llama 3 & \cellcolor{C5!70}$49.1\% \pm \text{\small17.5}$ & \cellcolor{C5!66}$56.7\% \pm \text{\small18.0}$ & \cellcolor{C5!65}$60.2\% \pm \text{\small18.1}$ \\
            \cellcolor{gray!20} & Llama 3.1 Base & \cellcolor{C5!23}$30.3\% \pm \text{\small9.7}$ & \cellcolor{C5!16}$30.4\% \pm \text{\small9.6}$ & \cellcolor{C5!13}$30.4\% \pm \text{\small9.6}$ \\
            \cellcolor{gray!20} & Llama 3.1 & \cellcolor{C5!59}$44.8\% \pm \text{\small13.6}$ & \cellcolor{C5!56}$51.5\% \pm \text{\small15.2}$ & \cellcolor{C5!57}$56.0\% \pm \text{\small16.9}$ \\
            \cellcolor{gray!20} & BLOOM & \cellcolor{C5!40}$36.9\% \pm \text{\small12.4}$ & \cellcolor{C5!48}$47.1\% \pm \text{\small13.7}$ & \cellcolor{C5!49}$51.5\% \pm \text{\small15.2}$ \\
            \cellcolor{gray!20} & Emo-BLOOM & \cellcolor{C5!49}$40.8\% \pm \text{\small12.2}$ & \cellcolor{C5!48}$47.0\% \pm \text{\small12.6}$ & \cellcolor{C5!47}$50.2\% \pm \text{\small12.5}$ \\
            \cellcolor{gray!20} & Qwen 2.5 Base & \cellcolor{C5!15}$27.0\% \pm \text{\small9.2}$ & \cellcolor{C5!40}$43.1\% \pm \text{\small14.1}$ & \cellcolor{C5!40}$46.0\% \pm \text{\small13.7}$ \\
            \cellcolor{gray!20} & Qwen 2.5 & \cellcolor{C5!62}$45.9\% \pm \text{\small15.7}$ & \cellcolor{C5!68}$57.8\% \pm \text{\small15.3}$ & \cellcolor{C5!67}$61.8\% \pm \text{\small16.2}$ \\
            \cellcolor{gray!20} & Qwen 3 Base & \cellcolor{C5!33}$34.2\% \pm \text{\small14.0}$ & \cellcolor{C5!31}$38.1\% \pm \text{\small17.2}$ & \cellcolor{C5!26}$38.3\% \pm \text{\small17.1}$ \\
            \cellcolor{gray!20} & Qwen 3 & \cellcolor{C5!62}$45.8\% \pm \text{\small13.9}$ & \cellcolor{C5!64}$55.8\% \pm \text{\small16.0}$ & \cellcolor{C5!60}$57.7\% \pm \text{\small16.4}$ \\
            \cellcolor{gray!20} & Mistral Base & \cellcolor{C5!27}$31.5\% \pm \text{\small11.6}$ & \cellcolor{C5!20}$32.5\% \pm \text{\small11.6}$ & \cellcolor{C5!17}$32.9\% \pm \text{\small11.9}$ \\
            \cellcolor{gray!20} & Mistral & \cellcolor{C5!52}$42.0\% \pm \text{\small13.3}$ & \cellcolor{C5!50}$48.3\% \pm \text{\small13.7}$ & \cellcolor{C5!50}$52.0\% \pm \text{\small14.8}$ \\
            \cellcolor{gray!20} & Ministral & \cellcolor{C5!61}$45.7\% \pm \text{\small15.9}$ & \cellcolor{C5!59}$53.1\% \pm \text{\small15.0}$ & \cellcolor{C5!59}$57.2\% \pm \text{\small14.4}$ \\
            \cellcolor{gray!20} & Ministral 3 Base & \cellcolor{C5!23}$30.3\% \pm \text{\small8.4}$ & \cellcolor{C5!17}$30.8\% \pm \text{\small8.5}$ & \cellcolor{C5!13}$30.9\% \pm \text{\small8.5}$ \\
            \multirow{-23}{*}{\cellcolor{gray!20} \makecell{\rotatebox{90}{English}}} & Ministral 3 & \cellcolor{C5!55}$43.3\% \pm \text{\small15.2}$ & \cellcolor{C5!60}$53.7\% \pm \text{\small16.9}$ & \cellcolor{C5!62}$58.4\% \pm \text{\small16.6}$ \\
            \midrule
            \cellcolor{gray!20} & Llama 2 & \cellcolor{C5!48}$39.6\% \pm \text{\small11.0}$ & \cellcolor{C5!36}$43.0\% \pm \text{\small11.2}$ & \cellcolor{C5!35}$43.7\% \pm \text{\small11.1}$ \\
            \cellcolor{gray!20} & EmoLlama 2 & \cellcolor{C5!45}$38.6\% \pm \text{\small10.2}$ & \cellcolor{C5!45}$47.1\% \pm \text{\small12.3}$ & \cellcolor{C5!45}$48.9\% \pm \text{\small13.4}$ \\
            \cellcolor{gray!20} & OLMo & \cellcolor{C5!40}$36.9\% \pm \text{\small11.0}$ & \cellcolor{C5!26}$38.3\% \pm \text{\small11.0}$ & \cellcolor{C5!25}$38.9\% \pm \text{\small11.0}$ \\
            \cellcolor{gray!20} & OLMo 2 & \cellcolor{C5!54}$41.4\% \pm \text{\small12.9}$ & \cellcolor{C5!57}$52.5\% \pm \text{\small14.0}$ & \cellcolor{C5!56}$54.4\% \pm \text{\small14.7}$ \\
            \cellcolor{gray!20} & OLMo 3 Base & \cellcolor{C5!20}$30.7\% \pm \text{\small8.0}$ & \cellcolor{C5!10}$30.9\% \pm \text{\small8.1}$ & \cellcolor{C5!10}$30.9\% \pm \text{\small8.1}$ \\
            \cellcolor{gray!20} & OLMo 3 & \cellcolor{C5!58}$42.9\% \pm \text{\small14.2}$ & \cellcolor{C5!50}$49.4\% \pm \text{\small13.4}$ & \cellcolor{C5!54}$53.3\% \pm \text{\small14.4}$ \\
            \cellcolor{gray!20} & Deepseek (Llama) & \cellcolor{C5!10}$27.2\% \pm \text{\small8.8}$ & \cellcolor{C5!10}$30.6\% \pm \text{\small8.1}$ & \cellcolor{C5!11}$31.5\% \pm \text{\small7.7}$ \\
            \cellcolor{gray!20} & Deepseek (Qwen) & \cellcolor{C5!14}$28.7\% \pm \text{\small8.2}$ & \cellcolor{C5!12}$31.6\% \pm \text{\small9.4}$ & \cellcolor{C5!14}$33.1\% \pm \text{\small9.4}$ \\
            \cellcolor{gray!20} & Llama 3 Base & \cellcolor{C5!44}$38.2\% \pm \text{\small9.8}$ & \cellcolor{C5!26}$38.5\% \pm \text{\small10.0}$ & \cellcolor{C5!25}$38.7\% \pm \text{\small10.2}$ \\
            \cellcolor{gray!20} & Llama 3 & \cellcolor{C5!70}$46.5\% \pm \text{\small18.1}$ & \cellcolor{C5!56}$52.2\% \pm \text{\small17.1}$ & \cellcolor{C5!61}$56.7\% \pm \text{\small18.4}$ \\
            \cellcolor{gray!20} & Llama 3.1 Base & \cellcolor{C5!49}$39.8\% \pm \text{\small10.6}$ & \cellcolor{C5!30}$40.0\% \pm \text{\small10.7}$ & \cellcolor{C5!28}$40.1\% \pm \text{\small10.6}$ \\
            \cellcolor{gray!20} & Llama 3.1 & \cellcolor{C5!63}$44.3\% \pm \text{\small16.8}$ & \cellcolor{C5!57}$52.5\% \pm \text{\small17.2}$ & \cellcolor{C5!61}$56.8\% \pm \text{\small18.3}$ \\
            \cellcolor{gray!20} & BLOOM & \cellcolor{C5!32}$34.3\% \pm \text{\small12.1}$ & \cellcolor{C5!30}$40.1\% \pm \text{\small13.2}$ & \cellcolor{C5!35}$43.7\% \pm \text{\small16.0}$ \\
            \cellcolor{gray!20} & Emo-BLOOM & \cellcolor{C5!33}$34.8\% \pm \text{\small12.1}$ & \cellcolor{C5!33}$41.4\% \pm \text{\small11.2}$ & \cellcolor{C5!34}$43.3\% \pm \text{\small11.2}$ \\
            \cellcolor{gray!20} & Qwen 2.5 Base & \cellcolor{C5!42}$37.5\% \pm \text{\small13.3}$ & \cellcolor{C5!39}$44.2\% \pm \text{\small12.4}$ & \cellcolor{C5!43}$47.7\% \pm \text{\small14.8}$ \\
            \cellcolor{gray!20} & Qwen 2.5 & \cellcolor{C5!52}$40.8\% \pm \text{\small14.0}$ & \cellcolor{C5!56}$52.0\% \pm \text{\small16.8}$ & \cellcolor{C5!70}$61.1\% \pm \text{\small18.7}$ \\
            \cellcolor{gray!20} & Qwen 3 Base & \cellcolor{C5!42}$37.7\% \pm \text{\small10.9}$ & \cellcolor{C5!27}$38.9\% \pm \text{\small11.0}$ & \cellcolor{C5!28}$40.2\% \pm \text{\small11.3}$ \\
            \cellcolor{gray!20} & Qwen 3 & \cellcolor{C5!51}$40.6\% \pm \text{\small15.5}$ & \cellcolor{C5!50}$49.5\% \pm \text{\small15.1}$ & \cellcolor{C5!60}$56.2\% \pm \text{\small17.4}$ \\
            \cellcolor{gray!20} & Mistral Base & \cellcolor{C5!21}$31.0\% \pm \text{\small7.0}$ & \cellcolor{C5!11}$31.5\% \pm \text{\small7.0}$ & \cellcolor{C5!11}$31.5\% \pm \text{\small7.0}$ \\
            \cellcolor{gray!20} & Mistral & \cellcolor{C5!59}$43.2\% \pm \text{\small12.9}$ & \cellcolor{C5!49}$49.0\% \pm \text{\small14.6}$ & \cellcolor{C5!49}$50.6\% \pm \text{\small14.5}$ \\
            \cellcolor{gray!20} & Ministral & \cellcolor{C5!56}$42.2\% \pm \text{\small13.3}$ & \cellcolor{C5!51}$50.0\% \pm \text{\small14.6}$ & \cellcolor{C5!52}$52.3\% \pm \text{\small13.8}$ \\
            \cellcolor{gray!20} & Ministral 3 Base & \cellcolor{C5!44}$38.3\% \pm \text{\small10.8}$ & \cellcolor{C5!26}$38.4\% \pm \text{\small10.8}$ & \cellcolor{C5!24}$38.4\% \pm \text{\small10.8}$ \\
            \multirow{-23}{*}{\cellcolor{gray!20} \makecell{\rotatebox{90}{Spanish}}}  & Ministral 3 & \cellcolor{C5!67}$45.7\% \pm \text{\small19.8}$ & \cellcolor{C5!70}$58.4\% \pm \text{\small20.3}$ & \cellcolor{C5!69}$60.9\% \pm \text{\small21.1}$ \\
            \bottomrule
        \end{tabular}
    }
    \caption{Full top-k similarity scores with $\pm$ 2SE's for models in the identification task. }
    \label{tab:full-ident-benchmark}
\end{table}
}

\newcommand{\condensedExpressionResults}{
\begin{table}[htbp]
    \centering
    \resizebox{0.95\linewidth}{!}{
        \begin{tabular}{ccccc}
            \toprule
            & \multirow{2}{*}{Model} & \multicolumn{3}{c}{Top-$k$ Similarity$\uparrow$} \\
            \cmidrule(lr){3-5}
            & & 1 & 3 & 5 \\
            \midrule
            \cellcolor{gray!20} & OLMo 3 Base & \cellcolor{C5!44}$42.5\% \pm \text{\small0.7}$ & \cellcolor{C5!32}$51.4\% \pm \text{\small0.7}$ & \cellcolor{C5!24}$55.6\% \pm \text{\small0.7}$ \\
            \cellcolor{gray!20} & OLMo 3 & \cellcolor{C5!70}$45.4\% \pm \text{\small0.6}$ & \cellcolor{C5!70}$55.3\% \pm \text{\small0.6}$ & \cellcolor{C5!70}$60.3\% \pm \text{\small0.6}$ \\
            \cellcolor{gray!20} & Llama 3.1 Base & \cellcolor{C5!46}$42.7\% \pm \text{\small0.6}$ & \cellcolor{C5!33}$51.5\% \pm \text{\small0.6}$ & \cellcolor{C5!23}$55.5\% \pm \text{\small0.6}$ \\
            \cellcolor{gray!20} & Llama 3.1 & \cellcolor{C5!10}$38.4\% \pm \text{\small0.7}$ & \cellcolor{C5!10}$49.1\% \pm \text{\small0.7}$ & \cellcolor{C5!10}$54.1\% \pm \text{\small0.7}$ \\
            \cellcolor{gray!20} & Qwen 3 Base & \cellcolor{C5!24}$40.1\% \pm \text{\small0.9}$ & \cellcolor{C5!22}$50.4\% \pm \text{\small0.9}$ & \cellcolor{C5!11}$54.3\% \pm \text{\small0.9}$ \\
            \cellcolor{gray!20} & Qwen 3 & \cellcolor{C5!51}$43.3\% \pm \text{\small0.6}$ & \cellcolor{C5!48}$53.1\% \pm \text{\small0.7}$ & \cellcolor{C5!37}$57.0\% \pm \text{\small0.7}$ \\
            \cellcolor{gray!20} & Ministral 3 Base & \cellcolor{C5!18}$39.4\% \pm \text{\small0.9}$ & \cellcolor{C5!21}$50.3\% \pm \text{\small0.9}$ & \cellcolor{C5!17}$54.9\% \pm \text{\small0.9}$ \\
            \multirow{-8}{*}{\cellcolor{gray!20}\makecell{\rotatebox{90}{English}}} & Ministral 3 & \cellcolor{C5!41}$42.1\% \pm \text{\small0.6}$ & \cellcolor{C5!39}$52.2\% \pm \text{\small0.7}$ & \cellcolor{C5!28}$56.1\% \pm \text{\small0.7}$ \\
            \midrule
            \cellcolor{gray!20} & OLMo 3 Base & \cellcolor{C5!28}$38.2\% \pm \text{\small0.9}$ & \cellcolor{C5!27}$48.5\% \pm \text{\small0.9}$ & \cellcolor{C5!25}$53.5\% \pm \text{\small0.9}$ \\
            \cellcolor{gray!20} & OLMo 3 & \cellcolor{C5!38}$40.2\% \pm \text{\small0.8}$ & \cellcolor{C5!38}$50.8\% \pm \text{\small0.8}$ & \cellcolor{C5!41}$56.7\% \pm \text{\small0.9}$ \\
            \cellcolor{gray!20} & Llama 3.1 Base & \cellcolor{C5!32}$38.9\% \pm \text{\small0.7}$ & \cellcolor{C5!24}$47.9\% \pm \text{\small0.7}$ & \cellcolor{C5!22}$52.8\% \pm \text{\small0.8}$ \\
            \cellcolor{gray!20} & Llama 3.1 & \cellcolor{C5!19}$36.3\% \pm \text{\small0.7}$ & \cellcolor{C5!17}$46.3\% \pm \text{\small0.8}$ & \cellcolor{C5!14}$51.4\% \pm \text{\small0.8}$ \\
            \cellcolor{gray!20} & Qwen 3 Base & \cellcolor{C5!62}$45.3\% \pm \text{\small1.2}$ & \cellcolor{C5!62}$56.0\% \pm \text{\small1.1}$ & \cellcolor{C5!63}$61.2\% \pm \text{\small1.1}$ \\
            \cellcolor{gray!20} & Qwen 3 & \cellcolor{C5!21}$36.7\% \pm \text{\small0.8}$ & \cellcolor{C5!28}$48.7\% \pm \text{\small0.9}$ & \cellcolor{C5!33}$55.2\% \pm \text{\small0.9}$ \\
            \cellcolor{gray!20} & Ministral 3 Base & \cellcolor{C5!70}$46.8\% \pm \text{\small1.2}$ & \cellcolor{C5!70}$57.6\% \pm \text{\small1.2}$ & \cellcolor{C5!70}$62.5\% \pm \text{\small1.1}$ \\
            \multirow{-8}{*}{\cellcolor{gray!20}\makecell{\rotatebox{90}{Spanish}}} & Ministral 3 & \cellcolor{C5!10}$34.2\% \pm \text{\small0.7}$ & \cellcolor{C5!10}$44.8\% \pm \text{\small0.8}$ & \cellcolor{C5!10}$50.4\% \pm \text{\small0.8}$ \\
            \bottomrule
        \end{tabular}
    }
    \caption{Top-k similarity scores with $\pm$ 2SE's for a subset of models in generating responses expressing a given affective state. Scores are shaded relative to those of other models for each value of $k$ (darker indicating better performance).}
    \label{tab:exp-benchmark}
    \vspace{-1em}
\end{table}
}

\newcommand{\fullExpressionResults}{
\begin{table}[htbp]
    \centering
    \resizebox{0.95\linewidth}{!}{
        \begin{tabular}{ccccc}
            \toprule
            & \multirow{2}{*}{Model} & \multicolumn{3}{c}{Top-$k$ Similarity$\uparrow$} \\
            \cmidrule(lr){3-5}
            & & 1 & 3 & 5 \\
            \midrule
            \cellcolor{gray!20} & Llama 2 & \cellcolor{C5!51}$42.0\% \pm \text{\small0.8}$ & \cellcolor{C5!54}$52.6\% \pm \text{\small0.8}$ & \cellcolor{C5!53}$57.3\% \pm \text{\small0.8}$ \\
            \cellcolor{gray!20} & EmoLlama 2 & \cellcolor{C5!42}$40.4\% \pm \text{\small0.7}$ & \cellcolor{C5!44}$51.0\% \pm \text{\small0.8}$ & \cellcolor{C5!43}$55.5\% \pm \text{\small0.8}$ \\
            \cellcolor{gray!20} & OLMo & \cellcolor{C5!50}$41.9\% \pm \text{\small0.7}$ & \cellcolor{C5!44}$51.0\% \pm \text{\small0.7}$ & \cellcolor{C5!41}$55.1\% \pm \text{\small0.7}$ \\
            \cellcolor{gray!20} & OLMo 2 & \cellcolor{C5!50}$41.8\% \pm \text{\small0.7}$ & \cellcolor{C5!52}$52.3\% \pm \text{\small0.7}$ & \cellcolor{C5!50}$56.7\% \pm \text{\small0.7}$ \\
            \cellcolor{gray!20} & OLMo 3 Base & \cellcolor{C5!53}$42.5\% \pm \text{\small0.7}$ & \cellcolor{C5!46}$51.4\% \pm \text{\small0.7}$ & \cellcolor{C5!44}$55.6\% \pm \text{\small0.7}$ \\
            \cellcolor{gray!20} & OLMo 3 & \cellcolor{C5!70}$45.4\% \pm \text{\small0.6}$ & \cellcolor{C5!70}$55.3\% \pm \text{\small0.6}$ & \cellcolor{C5!70}$60.3\% \pm \text{\small0.6}$ \\
            \cellcolor{gray!20} & Deepseek (Llama) & \cellcolor{C5!30}$38.2\% \pm \text{\small0.8}$ & \cellcolor{C5!30}$48.7\% \pm \text{\small0.8}$ & \cellcolor{C5!28}$52.7\% \pm \text{\small0.8}$ \\
            \cellcolor{gray!20} & Deepseek (Qwen) & \cellcolor{C5!10}$34.6\% \pm \text{\small0.7}$ & \cellcolor{C5!10}$45.3\% \pm \text{\small0.8}$ & \cellcolor{C5!10}$49.2\% \pm \text{\small0.8}$ \\
            \cellcolor{gray!20} & Llama 3 Base & \cellcolor{C5!54}$42.7\% \pm \text{\small0.6}$ & \cellcolor{C5!48}$51.7\% \pm \text{\small0.6}$ & \cellcolor{C5!44}$55.5\% \pm \text{\small0.6}$ \\
            \cellcolor{gray!20} & Llama 3 & \cellcolor{C5!38}$39.7\% \pm \text{\small0.7}$ & \cellcolor{C5!36}$49.7\% \pm \text{\small0.7}$ & \cellcolor{C5!38}$54.5\% \pm \text{\small0.7}$ \\
            \cellcolor{gray!20} & Llama 3.1 Base & \cellcolor{C5!54}$42.7\% \pm \text{\small0.6}$ & \cellcolor{C5!47}$51.5\% \pm \text{\small0.6}$ & \cellcolor{C5!44}$55.5\% \pm \text{\small0.6}$ \\
            \cellcolor{gray!20} & Llama 3.1 & \cellcolor{C5!31}$38.4\% \pm \text{\small0.7}$ & \cellcolor{C5!32}$49.1\% \pm \text{\small0.7}$ & \cellcolor{C5!36}$54.1\% \pm \text{\small0.7}$ \\
            \cellcolor{gray!20} & BLOOM & \cellcolor{C5!50}$41.9\% \pm \text{\small0.5}$ & \cellcolor{C5!37}$49.9\% \pm \text{\small0.5}$ & \cellcolor{C5!32}$53.4\% \pm \text{\small0.5}$ \\
            \cellcolor{gray!20} & Emo-BLOOM & \cellcolor{C5!47}$41.4\% \pm \text{\small0.4}$ & \cellcolor{C5!32}$49.1\% \pm \text{\small0.5}$ & \cellcolor{C5!28}$52.6\% \pm \text{\small0.5}$ \\
            \cellcolor{gray!20} & Qwen 2.5 Base & \cellcolor{C5!62}$44.0\% \pm \text{\small0.6}$ & \cellcolor{C5!54}$52.6\% \pm \text{\small0.6}$ & \cellcolor{C5!48}$56.3\% \pm \text{\small0.6}$ \\
            \cellcolor{gray!20} & Qwen 2.5 & \cellcolor{C5!52}$42.3\% \pm \text{\small0.6}$ & \cellcolor{C5!46}$51.4\% \pm \text{\small0.6}$ & \cellcolor{C5!43}$55.4\% \pm \text{\small0.6}$ \\
            \cellcolor{gray!20} & Qwen 3 Base & \cellcolor{C5!40}$40.1\% \pm \text{\small0.9}$ & \cellcolor{C5!40}$50.4\% \pm \text{\small0.9}$ & \cellcolor{C5!37}$54.3\% \pm \text{\small0.9}$ \\
            \cellcolor{gray!20} & Qwen 3 & \cellcolor{C5!58}$43.3\% \pm \text{\small0.6}$ & \cellcolor{C5!56}$53.1\% \pm \text{\small0.7}$ & \cellcolor{C5!51}$57.0\% \pm \text{\small0.7}$ \\
            \cellcolor{gray!20} & Mistral Base & \cellcolor{C5!41}$40.3\% \pm \text{\small0.8}$ & \cellcolor{C5!37}$49.9\% \pm \text{\small0.9}$ & \cellcolor{C5!37}$54.4\% \pm \text{\small0.9}$ \\
            \cellcolor{gray!20} & Mistral & \cellcolor{C5!43}$40.6\% \pm \text{\small0.7}$ & \cellcolor{C5!35}$49.5\% \pm \text{\small0.6}$ & \cellcolor{C5!29}$52.9\% \pm \text{\small0.6}$ \\
            \cellcolor{gray!20} & Ministral & \cellcolor{C5!50}$42.0\% \pm \text{\small0.7}$ & \cellcolor{C5!50}$52.1\% \pm \text{\small0.7}$ & \cellcolor{C5!50}$56.6\% \pm \text{\small0.7}$ \\
            \cellcolor{gray!20} & Ministral 3 Base & \cellcolor{C5!36}$39.4\% \pm \text{\small0.9}$ & \cellcolor{C5!40}$50.3\% \pm \text{\small0.9}$ & \cellcolor{C5!41}$54.9\% \pm \text{\small0.9}$ \\
            \multirow{-23}{*}{\cellcolor{gray!20}\makecell{\rotatebox{90}{English}}} & Ministral 3 & \cellcolor{C5!51}$42.1\% \pm \text{\small0.6}$ & \cellcolor{C5!51}$52.2\% \pm \text{\small0.7}$ & \cellcolor{C5!47}$56.1\% \pm \text{\small0.7}$ \\
            \midrule
            \cellcolor{gray!20} & Llama 2 & \cellcolor{C5!18}$36.0\% \pm \text{\small0.6}$ & \cellcolor{C5!10}$44.7\% \pm \text{\small0.7}$ & \cellcolor{C5!10}$49.6\% \pm \text{\small0.7}$ \\
            \cellcolor{gray!20} & EmoLlama 2 & \cellcolor{C5!22}$36.9\% \pm \text{\small0.6}$ & \cellcolor{C5!15}$45.8\% \pm \text{\small0.7}$ & \cellcolor{C5!14}$50.6\% \pm \text{\small0.8}$ \\
            \cellcolor{gray!20} & OLMo & \cellcolor{C5!28}$38.1\% \pm \text{\small0.8}$ & \cellcolor{C5!27}$48.4\% \pm \text{\small0.9}$ & \cellcolor{C5!29}$53.8\% \pm \text{\small0.9}$ \\
            \cellcolor{gray!20} & OLMo 2 & \cellcolor{C5!28}$38.2\% \pm \text{\small0.7}$ & \cellcolor{C5!22}$47.5\% \pm \text{\small0.7}$ & \cellcolor{C5!23}$52.5\% \pm \text{\small0.8}$ \\
            \cellcolor{gray!20} & OLMo 3 Base & \cellcolor{C5!28}$38.2\% \pm \text{\small0.9}$ & \cellcolor{C5!27}$48.5\% \pm \text{\small0.9}$ & \cellcolor{C5!28}$53.5\% \pm \text{\small0.9}$ \\
            \cellcolor{gray!20} & OLMo 3 & \cellcolor{C5!38}$40.2\% \pm \text{\small0.8}$ & \cellcolor{C5!38}$50.8\% \pm \text{\small0.8}$ & \cellcolor{C5!43}$56.7\% \pm \text{\small0.9}$ \\
            \cellcolor{gray!20} & Deepseek (Llama) & \cellcolor{C5!55}$43.9\% \pm \text{\small1.1}$ & \cellcolor{C5!57}$54.9\% \pm \text{\small1.1}$ & \cellcolor{C5!58}$59.9\% \pm \text{\small1.1}$ \\
            \cellcolor{gray!20} & Deepseek (Qwen) & \cellcolor{C5!16}$35.7\% \pm \text{\small0.8}$ & \cellcolor{C5!12}$45.3\% \pm \text{\small0.9}$ & \cellcolor{C5!10}$49.6\% \pm \text{\small0.9}$ \\
            \cellcolor{gray!20} & Llama 3 Base & \cellcolor{C5!35}$39.6\% \pm \text{\small0.7}$ & \cellcolor{C5!25}$48.0\% \pm \text{\small0.7}$ & \cellcolor{C5!23}$52.4\% \pm \text{\small0.8}$ \\
            \cellcolor{gray!20} & Llama 3 & \cellcolor{C5!17}$35.8\% \pm \text{\small0.7}$ & \cellcolor{C5!17}$46.2\% \pm \text{\small0.8}$ & \cellcolor{C5!17}$51.3\% \pm \text{\small0.8}$ \\
            \cellcolor{gray!20} & Llama 3.1 Base & \cellcolor{C5!32}$38.9\% \pm \text{\small0.7}$ & \cellcolor{C5!25}$47.9\% \pm \text{\small0.7}$ & \cellcolor{C5!25}$52.8\% \pm \text{\small0.8}$ \\
            \cellcolor{gray!20} & Llama 3.1 & \cellcolor{C5!19}$36.3\% \pm \text{\small0.7}$ & \cellcolor{C5!17}$46.3\% \pm \text{\small0.8}$ & \cellcolor{C5!18}$51.4\% \pm \text{\small0.8}$ \\
            \cellcolor{gray!20} & BLOOM & \cellcolor{C5!37}$40.1\% \pm \text{\small0.6}$ & \cellcolor{C5!20}$47.0\% \pm \text{\small0.6}$ & \cellcolor{C5!16}$51.0\% \pm \text{\small0.6}$ \\
            \cellcolor{gray!20} & Emo-BLOOM & \cellcolor{C5!36}$39.7\% \pm \text{\small0.5}$ & \cellcolor{C5!19}$46.8\% \pm \text{\small0.6}$ & \cellcolor{C5!14}$50.5\% \pm \text{\small0.6}$ \\
            \cellcolor{gray!20} & Qwen 2.5 Base & \cellcolor{C5!29}$38.3\% \pm \text{\small0.7}$ & \cellcolor{C5!21}$47.2\% \pm \text{\small0.7}$ & \cellcolor{C5!22}$52.2\% \pm \text{\small0.8}$ \\
            \cellcolor{gray!20} & Qwen 2.5 & \cellcolor{C5!28}$38.2\% \pm \text{\small0.7}$ & \cellcolor{C5!20}$46.8\% \pm \text{\small0.7}$ & \cellcolor{C5!21}$52.1\% \pm \text{\small0.8}$ \\
            \cellcolor{gray!20} & Qwen 3 Base & \cellcolor{C5!62}$45.3\% \pm \text{\small1.2}$ & \cellcolor{C5!62}$56.0\% \pm \text{\small1.1}$ & \cellcolor{C5!63}$61.2\% \pm \text{\small1.1}$ \\
            \cellcolor{gray!20} & Qwen 3 & \cellcolor{C5!21}$36.7\% \pm \text{\small0.8}$ & \cellcolor{C5!28}$48.7\% \pm \text{\small0.9}$ & \cellcolor{C5!36}$55.2\% \pm \text{\small0.9}$ \\
            \cellcolor{gray!20} & Mistral Base & \cellcolor{C5!45}$41.6\% \pm \text{\small0.9}$ & \cellcolor{C5!38}$50.8\% \pm \text{\small1.0}$ & \cellcolor{C5!36}$55.3\% \pm \text{\small1.0}$ \\
            \cellcolor{gray!20} & Mistral & \cellcolor{C5!31}$38.7\% \pm \text{\small0.7}$ & \cellcolor{C5!29}$48.9\% \pm \text{\small0.8}$ & \cellcolor{C5!30}$54.0\% \pm \text{\small0.8}$ \\
            \cellcolor{gray!20} & Ministral & \cellcolor{C5!38}$40.1\% \pm \text{\small0.7}$ & \cellcolor{C5!27}$48.5\% \pm \text{\small0.7}$ & \cellcolor{C5!28}$53.5\% \pm \text{\small0.8}$ \\
            \cellcolor{gray!20} & Ministral 3 Base & \cellcolor{C5!70}$46.8\% \pm \text{\small1.2}$ & \cellcolor{C5!70}$57.6\% \pm \text{\small1.2}$ & \cellcolor{C5!70}$62.5\% \pm \text{\small1.1}$ \\
            \multirow{-23}{*}{\cellcolor{gray!20}\makecell{\rotatebox{90}{Spanish}}} & Ministral 3 & \cellcolor{C5!10}$34.2\% \pm \text{\small0.7}$ & \cellcolor{C5!10}$44.8\% \pm \text{\small0.8}$ & \cellcolor{C5!13}$50.4\% \pm \text{\small0.8}$ \\
            \bottomrule
        \end{tabular}
    }
    \caption{Full top-k similarity scores with $\pm$ 2SE's for all models in generating responses expressing a given affective state. }
    \label{tab:full-exp-benchmark}
\end{table}
}

\newcommand\rankingOne{
    \begin{tabular}{cccc}
        \toprule
        \multicolumn{4}{c}{\textbf{OLMo 3}} \\
        \toprule
        \multicolumn{2}{c}{English} & \multicolumn{2}{c}{Spanish} \\
        \cmidrule(lr){1-2} \cmidrule(lr){3-4}
        Term & Rank & Term & Rank \\
        \midrule
        \textbf{abandoned} & 3.3 & \textbf{frustrada} & 5.7 \\
        dizzy & 9.0 & \textbf{afortunada} & 8.3 \\
        panicked & 9.0 & curioso & 8.3 \\
        \textbf{lonely} & 9.3 & \textit{triste} & 10.0 \\
        appalled & 10.0 & \textit{deprimida} & 11.0 \\
        \midrule
        \textit{reluctant} & 30.7 & \textbf{rodeado} & 34.7 \\
        \textbf{aloof} & 31.3 & contra & 37.0 \\
        \textit{downtrodden} & 31.3 & \textit{convencido} & 37.3 \\
        wide & 33.3 & obligado & 38.3 \\
        jumpy & 35.3 & \textbf{trabado} & 39.0 \\
        \bottomrule
    \end{tabular}
}

\newcommand\rankingTwo{
    \begin{tabular}{cccc}
        \toprule
        \multicolumn{4}{c}{\textbf{Llama 3.1}} \\
        \toprule
        \multicolumn{2}{c}{English} & \multicolumn{2}{c}{Spanish} \\
        \cmidrule(lr){1-2} \cmidrule(lr){3-4}
        Term & Rank & Term & Rank \\
        \midrule
        \textbf{lonely} & 5.0 & \textbf{afortunada} & 4.7 \\
        \textit{upset} & 6.3 & \textbf{frustrada} & 5.3 \\
        \textbf{abandoned} & 7.7 & nerviosa & 9.0 \\
        paralyzed & 10.0 & devastado & 9.0 \\
        unnoticed & 13.7 & \textit{deprimida} & 9.3 \\
        \midrule
        \textbf{aloof} & 29.3 & dormido & 35.7 \\
        done & 29.7 & \textbf{rodeado} & 35.7 \\
        \textit{wide} & 29.7 & fea & 36.7 \\
        \textit{jittery} & 31.0 & \textbf{trabado} & 38.0 \\
        unbalanced & 36.3 & echo & 39.3 \\
        \bottomrule
    \end{tabular}
}

\newcommand\rankingThree{
    \begin{tabular}{cccc}
        \toprule
        \multicolumn{4}{c}{\textbf{Qwen 3}} \\
        \toprule
        \multicolumn{2}{c}{English} & \multicolumn{2}{c}{Spanish} \\
        \cmidrule(lr){1-2} \cmidrule(lr){3-4}
        Term & Rank & Term & Rank \\
        \midrule
        \textbf{abandoned} & 3.0 & \textbf{afortunada} & 6.7 \\
        \textit{upset} & 4.7 & \textit{deprimida} & 9.7 \\
        \textit{nervous} & 7.7 & \textit{ignorada} & 10.3 \\
        aggressive & 7.7 & \textit{triste} & 10.3 \\
        \textbf{lonely} & 7.7 & \textbf{frustrada} & 11.7 \\
        \midrule
        inexperienced & 31.7 & querido & 34.3 \\
        \textit{downtrodden} & 32.0 & contra & 34.7 \\
        \textit{reluctant} & 32.3 & \textit{convencido} & 35.7 \\
        \textit{jittery} & 33.3 & \textbf{trabado} & 38.0 \\
        \textbf{aloof} & 35.3 & \textbf{rodeado} & 41.0 \\
        \bottomrule
    \end{tabular}
}

\newcommand\rankingFour{
    \begin{tabular}{cccc}
        \toprule
        \multicolumn{4}{c}{\textbf{Ministral 3}} \\
        \toprule
        \multicolumn{2}{c}{English} & \multicolumn{2}{c}{Spanish} \\
        \cmidrule(lr){1-2} \cmidrule(lr){3-4}
        Term & Rank & Term & Rank \\
        \midrule
        \textbf{abandoned} & 4.0 & \textbf{frustrada} & 7.7 \\
        \textbf{lonely} & 5.7 & \textbf{afortunada} & 9.7 \\
        supported & 9.3 & incomoda & 11.3 \\
        deep & 10.3 & \textit{ignorada} & 12.3 \\
        \textit{nervous} & 10.7 & mejor & 12.3 \\
        \midrule
        \textbf{aloof} & 30.3 & contra & 36.0 \\
        \textit{downtrodden} & 30.7 & mareado & 37.7 \\
        \textit{jittery} & 31.7 & \textbf{trabado} & 39.7 \\
        dizzy & 31.7 & \textit{convencido} & 40.0 \\
        rested & 34.3 & \textbf{rodeado} & 40.0 \\
        \bottomrule
    \end{tabular}
}

\newcommand\rankingTables{

    \begin{table*}[htbp]
        \centering
        
        \begin{subtable}[t]{0.49\textwidth}
            \centering
            \resizebox{\textwidth}{!}{
                \rankingOne
            }
        \end{subtable}
        \hfill
        \begin{subtable}[t]{0.48\textwidth}
            \centering
            \resizebox{\textwidth}{!}{
                \rankingTwo
            }
        \end{subtable}
        \begin{subtable}[t]{0.49\textwidth}
            \centering
            \resizebox{\textwidth}{!}{
                \rankingThree
            }
        \end{subtable}
        \hfill
        \begin{subtable}[t]{0.49\textwidth}
            \centering
            \resizebox{\textwidth}{!}{
                \rankingFour
            }
        \end{subtable}
        \caption{Top and bottom-5 affective states by average performance rank across the three tasks. \textbf{Bolded} terms appear in the top/bottom ranks for all four models, while \textit{italicized} terms appear for two or three. 
        }
        \label{tab:rank-exs}
    \end{table*}
}

\newcommand\rankingTablesV{
    \begin{table}[t]
        \centering
        
        \begin{subtable}[t]{0.95\linewidth}
            \centering
            \resizebox{\textwidth}{!}{
                \rankingOne
            }
        \end{subtable}
        \hfill
        \begin{subtable}[t]{0.95\linewidth}
            \centering
            \resizebox{\textwidth}{!}{
                \rankingTwo
            }
        \end{subtable}
        \hfill
        \begin{subtable}[t]{0.95\linewidth}
            \centering
            \resizebox{\textwidth}{!}{
                \rankingThree
            }
        \end{subtable}
        
        \caption{Top and bottom-3 affective states by average performance ranking across the three tasks. \textbf{Bolded} terms appear in the top/bottom ranks for all three models, while \textit{italicized} terms appear for two.}
        \label{tab:rank-exs}
    \end{table}

}

\newcommand\condensedEnCoefficients{

    \begin{table*}[]
        \centering
        \resizebox{\linewidth}{!}{
            \begin{tabular}{ccccccccc}
                \toprule
                & OLMo 3 Base & OLMo 3 & Llama 3.1 Base & Llama 3.1 & Qwen 3 Base & Qwen 3 & Ministral 3 Base & Ministral 3 \\
                \midrule
                Freq & $0.50$ & $0.06$ & $0.78$ & $0.12$ & $1.01^{*}$ & $0.43$ & $0.31$ & $0.41$ \\
                Valence & $-0.54$ & $-0.15$ & $-0.18$ & $-0.03$ & $-0.84^{\dagger}$ & $-0.52$ & $-0.74$ & $-0.43$ \\
                Arousal & $0.51$ & $0.53$ & $0.38$ & $0.28$ & $0.86^{*}$ & $0.44$ & $0.20$ & $0.11$ \\
                Conc & $-0.29$ & $0.23$ & $0.08$ & $0.27$ & $-0.27$ & $-0.08$ & $-0.23$ & $0.25$ \\
                AoA & $-1.16^{*}$ & $-0.92^{\dagger}$ & $-1.00^{\dagger}$ & $-1.24^{*}$ & $-0.76$ & $-1.25^{*}$ & $-0.79$ & $-1.07^{*}$ \\
                \bottomrule
            \end{tabular}
        }
        \caption{Coefficients of the Ordinal Regression Model predicting affective state rank in English. Superscripts indicate significant coefficients, $^{**} p \leq .01$, $^{*} p \leq .05$, $^{\dagger} p \leq .075$.}
        \label{tab:en-wev-coefs}
    \end{table*}   
    
}

\newcommand\condensedEsCoefficients{

    \begin{table*}[]
        \centering
        \resizebox{\linewidth}{!}{
            \begin{tabular}{ccccccccc}
                \toprule
                & OLMo 3 Base & OLMo 3 & Llama 3.1 Base & Llama 3.1 & Qwen 3 Base & Qwen 3 & Ministral 3 Base & Ministral 3 \\
                \midrule
                Freq & $2.46$ & $-0.77$ & $0.24$ & $-1.65$ & $-1.93$ & $0.14$ & $0.13$ & $0.90$ \\
                Valence & $0.60$ & $1.16$ & $-0.30$ & $0.90$ & $0.22$ & $2.24$ & $1.02$ & $1.28$ \\
                Arousal & $5.30$ & $1.89$ & $1.25$ & $2.16$ & $0.46$ & $5.85$ & $4.65$ & $3.75$ \\
                Aoa & $5.16$ & $0.90$ & $1.63$ & $-0.14$ & $-0.03$ & $1.08$ & $3.03$ & $1.77$ \\
                \bottomrule
            \end{tabular}
        }
        \caption{Coefficients of the Ordinal Regression Model predicting affective state rank in Spanish. Superscripts indicate significant coefficients, $^{**} p \leq .01$, $^{*} p \leq .05$, $^{\dagger} p \leq .075$.}
        \label{tab:es-wev-coefs}
    \end{table*}   
    
}

\newcommand\condensedCoefficients{
    \begin{table*}[htbp]
        \centering
        \begin{subtable}{0.95\linewidth}
        \resizebox{\linewidth}{!}{
            \begin{tabular}{ccccccccc}
                \toprule
                \multicolumn{9}{c}{English} \\
                \toprule
                & OLMo 3 Base & OLMo 3 & Llama 3.1 Base & Llama 3.1 & Qwen 3 Base & Qwen 3 & Ministral 3 Base & Ministral 3 \\
                \midrule
                Freq & $0.62$ & $0.02$ & $0.91$ & $0.21$ & $1.11^{*}$ & $0.43$ & $0.49$ & $0.47$ \\
                Valence & $-0.56$ & $-0.14$ & $-0.20$ & $-0.06$ & $-0.84^{~}$ & $-0.51$ & $-0.80$ & $-0.44$ \\
                Arousal & $0.51$ & $0.51$ & $0.37$ & $0.30$ & $0.83^{*}$ & $0.42$ & $0.21$ & $0.10$ \\
                Conc & $-0.28$ & $0.23$ & $0.09$ & $0.28$ & $-0.25$ & $-0.07$ & $-0.22$ & $0.26$ \\
                AoA & $-1.13^{*}$ & $-0.95^{*}$ & $-1.00^{~}$ & $-1.20^{*}$ & $-0.77$ & $-1.28^{*}$ & $-0.73$ & $-1.06^{*}$ \\
                \bottomrule
            \end{tabular}
            }
        \end{subtable}
        \vspace{2em}
        \begin{subtable}{0.95\linewidth}
        \resizebox{\linewidth}{!}{
            \begin{tabular}{ccccccccc}
                \toprule
                \multicolumn{9}{c}{Spanish} \\
                \toprule
                & OLMo 3 Base & OLMo 3 & Llama 3.1 Base & Llama 3.1 & Qwen 3 Base & Qwen 3 & Ministral 3 Base & Ministral 3 \\
                \midrule
                Freq & $1.20$ & $-1.72$ & $0.01$ & $-2.76$ & $-2.53$ & $-0.52$ & $-0.48$ & $-0.04$ \\
                Valence & $0.39$ & $1.56$ & $-0.28$ & $1.57$ & $0.85$ & $2.29$ & $1.21$ & $1.23$ \\
                Arousal & $2.60$ & $3.02$ & $1.06$ & $4.44$ & $2.88$ & $5.73^{\dagger}$ & $4.80$ & $2.74$ \\
                AoA & $3.67$ & $0.28$ & $1.43$ & $-0.72$ & $-0.20$ & $0.58$ & $2.59$ & $0.91$ \\
                \bottomrule
            \end{tabular}
            }
        \end{subtable}
        \vspace{-2em}
        \caption{Coefficients of the Ordinal Regression Model predicting affective state rank in English (top) and Spanish (bottom). Superscripts indicate significant coefficients, $^{**} p \leq .01$, $^{*} p \leq .05$, $^{\dagger} p \leq .075$.}
        \label{tab:wev-coefs}
        \vspace{-1em}
    \end{table*}
}

\newcommand\condensedCoefficientsLarge{

    \begin{table*}[htbp]
        \centering
        \resizebox{\linewidth}{!}{
            \begin{tabular}{cccccccccc}
                \toprule
                & & OLMo 3 Base & OLMo 3 & Llama 3.1 Base & Llama 3.1 & Qwen 3 Base & Qwen 3 & Ministral 3 Base & Ministral 3 \\
                \midrule
                \cellcolor{gray!20} & Freq & $0.63^{*}$ & $0.42^{\dagger}$ & $0.67^{**}$ & $0.49^{\dagger}$ & $0.43^{\dagger}$ & $0.57^{*}$ & $0.11$ & $0.48^{\dagger}$ \\
                \cellcolor{gray!20} & Valence & $-0.18$ & $-0.04$ & $-0.18$ & $-0.09$ & $-0.13$ & $-0.00$ & $-0.15$ & $0.11$ \\
                \cellcolor{gray!20} & Arousal & $0.08$ & $-0.03$ & $-0.04$ & $-0.07$ & $0.15$ & $-0.14$ & $-0.14$ & $-0.11$ \\
                \cellcolor{gray!20} & Conc & $0.19$ & $0.21$ & $0.07$ & $0.09$ & $0.15$ & $0.35^{*}$ & $-0.06$ & $0.22$ \\
                \multirow{-5}{*}{\cellcolor{gray!20}\makecell{\rotatebox{90}{English}}} & AoA & $-0.28$ & $0.01$ & $-0.51^{*}$ & $-0.28$ & $0.08$ & $0.03$ & $0.07$ & $-0.35$ \\
                \midrule
                \cellcolor{gray!20} & Freq & $-0.51^{**}$ & $-0.13$ & $-0.53^{**}$ & $-0.48^{*}$ & $-0.90^{**}$ & $-0.03$ & $-0.94^{**}$ & $0.18$ \\
                \cellcolor{gray!20} & Valence & $-0.10$ & $0.29$ & $0.02$ & $0.30$ & $0.07$ & $0.38^{\dagger}$ & $-0.01$ & $0.05$ \\
                \cellcolor{gray!20} & Arousal & $0.24$ & $-0.16$ & $0.05$ & $0.14$ & $0.36^{\dagger}$ & $-0.01$ & $0.42^{*}$ & $-0.01$ \\
                \multirow{-4}{*}{\cellcolor{gray!20}\makecell{\rotatebox{90}{Spanish}}} & AoA & $0.22$ & $0.53^{**}$ & $0.42^{*}$ & $0.35^{\dagger}$ & $0.05$ & $0.36^{\dagger}$ & $0.26$ & $0.27$ \\
                \bottomrule
            \end{tabular}
            }
        \caption{Coefficients of the Ordinal Regression Model predicting affective state rank in English (top) and Spanish (bottom). Superscripts indicate significant coefficients, $^{**} p \leq .01$, $^{*} p \leq .05$, $^{\dagger} p \leq .075$.}
        \label{tab:wev-coefs}
        \vspace{-1em}
    \end{table*}
}

\newcommand\condensedCoefficientsNoTF{

    \begin{table*}[htbp]
        \centering
        \resizebox{\linewidth}{!}{
            \begin{tabular}{cccccccccc}
                \toprule
                & & OLMo 3 Base & OLMo 3 & Llama 3.1 Base & Llama 3.1 & Qwen 3 Base & Qwen 3 & Ministral 3 Base & Ministral 3 \\
                \midrule
                \cellcolor{gray!20} & Valence & $-0.03$ & $0.06$ & $-0.00$ & $0.02$ & $-0.03$ & $0.15$ & $-0.12$ & $0.22$ \\
                \cellcolor{gray!20} & Arousal & $-0.06$ & $-0.12$ & $-0.18$ & $-0.18$ & $0.07$ & $-0.26$ & $-0.17$ & $-0.23$ \\
                \cellcolor{gray!20} & Conc & $0.19$ & $0.21$ & $0.08$ & $0.10$ & $0.16$ & $0.35^{*}$ & $-0.05$ & $0.23$ \\
                \multirow{-4}{*}{\cellcolor{gray!20}\makecell{\rotatebox{90}{English}}} & AoA & $-0.64^{**}$ & $-0.23$ & $-0.87^{**}$ & $-0.55^{**}$ & $-0.17$ & $-0.28^{\dagger}$ & $0.00$ & $-0.62^{**}$ \\
                \midrule
                \cellcolor{gray!20} & Valence & $-0.23$ & $0.24$ & $-0.13$ & $0.17$ & $-0.21$ & $0.37^{\dagger}$ & $-0.26$ & $0.11$ \\
                \cellcolor{gray!20} & Arousal & $0.27$ & $-0.14$ & $0.11$ & $0.20$ & $0.45^{*}$ & $-0.01$ & $0.45^{*}$ & $-0.04$ \\
                \multirow{-3}{*}{\cellcolor{gray!20}\makecell{\rotatebox{90}{Spanish}}} & AoA & $0.42^{*}$ & $0.58^{**}$ & $0.60^{**}$ & $0.50^{**}$ & $0.36^{\dagger}$ & $0.37^{*}$ & $0.58^{**}$ & $0.20$ \\
                \bottomrule
            \end{tabular}
            }
        \caption{Coefficients of the Ordinal Regression Model predicting affective state rank in English (top) and Spanish (bottom) excluding term frequency from the exogenous variables. Superscripts indicate significant coefficients, $^{**} p \leq .01$, $^{*} p \leq .05$, $^{\dagger} p \leq .075$.}
        \label{tab:wev-coefs-no-tf}
        \vspace{-1em}
    \end{table*}
}

\newcommand\condensedCoefficientsNoAoA{

    \begin{table*}[htbp]
        \centering
        \resizebox{\linewidth}{!}{
            \begin{tabular}{cccccccccc}
                \toprule
                & & OLMo 3 Base & OLMo 3 & Llama 3.1 Base & Llama 3.1 & Qwen 3 Base & Qwen 3 & Ministral 3 Base & Ministral 3 \\
                \midrule
                \cellcolor{gray!20} & Freq & $0.83^{**}$ & $0.42^{*}$ & $1.00^{**}$ & $0.67^{**}$ & $0.38^{*}$ & $0.54^{**}$ & $0.06$ & $0.73^{**}$ \\
                \cellcolor{gray!20} & Valence & $-0.23$ & $-0.03$ & $-0.26$ & $-0.13$ & $-0.12$ & $0.00$ & $-0.14$ & $0.06$ \\
                \cellcolor{gray!20} & Arousal & $0.11$ & $-0.03$ & $0.03$ & $-0.04$ & $0.15$ & $-0.14$ & $-0.15$ & $-0.06$ \\
                \multirow{-4}{*}{\cellcolor{gray!20}\makecell{\rotatebox{90}{English}}} & Conc & $0.23$ & $0.21$ & $0.13$ & $0.14$ & $0.14$ & $0.35^{*}$ & $-0.07$ & $0.27$ \\
                \midrule
                \cellcolor{gray!20} & Freq & $-0.58^{**}$ & $-0.28^{\dagger}$ & $-0.66^{**}$ & $-0.58^{**}$ & $-0.92^{**}$ & $-0.15$ & $-1.03^{**}$ & $0.09$ \\
                \cellcolor{gray!20} & Valence & $-0.10$ & $0.31$ & $0.03$ & $0.29$ & $0.07$ & $0.39^{\dagger}$ & $0.01$ & $0.07$ \\
                \multirow{-3}{*}{\cellcolor{gray!20}\makecell{\rotatebox{90}{Spanish}}} & Arousal & $0.29$ & $-0.04$ & $0.13$ & $0.22$ & $0.37^{\dagger}$ & $0.07$ & $0.46^{*}$ & $0.05$ \\
                \bottomrule
            \end{tabular}
            }
        \caption{Coefficients of the Ordinal Regression Model predicting affective state rank in English (top) and Spanish (bottom) excluding Age of Acquisition from the exogenous variables. Superscripts indicate significant coefficients, $^{**} p \leq .01$, $^{*} p \leq .05$, $^{\dagger} p \leq .075$.}
        \label{tab:wev-coefs-no-aoa}
        \vspace{-1em}
    \end{table*}
}

\newcommand\rsquaredAnalysis{
    \begin{table}[htbp]
        \begin{subtable}{0.95\linewidth}
            \centering
            \begin{tabular}{cccc}
                \toprule
                \multicolumn{4}{c}{English} \\
                \toprule
                Model & Full & No Freq & No AoA \\
                \midrule
                OLMo 3 Base & 0.076 & 0.057 & 0.071 \\
                OLMo 3 & 0.027 & 0.018 & 0.027 \\
                Llama 3.1 Base & 0.106 & 0.085 & 0.092 \\
                Llama 3.1 & 0.055 & 0.044 & 0.050 \\
                Qwen 3 Base & 0.017 & 0.008 & 0.017 \\
                Qwen 3 & 0.059 & 0.045 & 0.059 \\
                Ministral 3 Base & 0.005 & 0.004 & 0.004 \\
                Ministral 3 & 0.084 & 0.074 & 0.077 \\
                \bottomrule
            \end{tabular}
            \caption{Caption}
            \label{tab:placeholder}
        \end{subtable}
        \vspace{2em}
        \begin{subtable}{0.95\linewidth}
            \centering
            \begin{tabular}{cccc}
                \toprule
                \multicolumn{4}{c}{Spanish} \\
                \toprule
                Model & Full & No Freq & No AoA \\
                \midrule
                OLMo 3 Base & 0.076 & 0.057 & 0.071 \\
                OLMo 3 & 0.027 & 0.018 & 0.027 \\
                Llama 3.1 Base & 0.106 & 0.085 & 0.092 \\
                Llama 3.1 & 0.055 & 0.044 & 0.050 \\
                Qwen 3 Base & 0.017 & 0.008 & 0.017 \\
                Qwen 3 & 0.059 & 0.045 & 0.059 \\
                Ministral 3 Base & 0.005 & 0.004 & 0.004 \\
                Ministral 3 & 0.084 & 0.074 & 0.077 \\
                \bottomrule
            \end{tabular}
            \caption{Caption}
            \label{tab:placeholder}
        \end{subtable}
    \end{table}
}

\newcommand\taskCorrelations{
    \begin{table}[htbp]
        \centering
            \begin{tabular}{cccc}
                \toprule
                Model & I$\times$V & I$\times$E & V$\times$E \\
                \midrule
                \multicolumn{4}{c}{English}\\
                \midrule
                OLMo 3 & $0.11$ & $0.39^{*}$ & $-0.01$ \\
                Llama 3.1 & $0.04$ & $0.12$ & $-0.07$ \\
                Qwen 3 & $0.08$ & $0.40^{**}$ & $0.14$ \\
                Ministral & $-0.06$ & $0.45^{**}$ & $-0.08$ \\
                \midrule
                \multicolumn{4}{c}{Spanish}\\
                \midrule
                OLMo 3 & $0.10$ & $0.15$ & $0.18$ \\
                Llama 3.1 & $0.29^{\dagger}$ & $0.30^{*}$ & $0.47^{**}$ \\
                Qwen 3 & $0.06$ & $0.23$ & $0.13$ \\
                Ministral & $0.14$ & $0.09$ & $0.35^{*}$ \\
                \bottomrule
            \end{tabular}
        \caption{Affective state-level Spearman's correlations between performance measures in the \underline{\textbf{I}}dentification, \underline{\textbf{V}}erification, \underline{\textbf{E}}xpression. Superscripts indicate significant correlations, $^{**} p \leq .01$, $^{*} p\leq.05$, $^{\dagger} p\leq.075$.}
        \label{tab:task-corrs}
        \vspace{-1em}
    \end{table}
}

\newcommand{\modelDetails}{
    \begin{table*}[htbp]
        \centering
        \resizebox{\linewidth}{!}{
            \begin{tabular}{ccccc}
                \toprule
                Model & Checkpoint & Size & \multicolumn{2}{c}{Training} \\
                \cmidrule(lr){4-5}
                & & & Instruct & Emotion \\
                \midrule
                OLMo \cite{groeneveld-olmo}                         & \texttt{OLMo-7B-Instruct-hf}          & 7B  & \Checkmark & \\
                \midrule
                OLMo-2 \cite{walsh-olmo2}                       & \texttt{OLMo-2-1124-7B-Instruct}      & 7B  & \Checkmark & \\
                \midrule
                \multirow{2}{*}{OLMo-3 \cite{olmo-olmo3}}      & \texttt{allenai/Olmo-3-1025-7B}       & 7B  &            & \\
                                                     & \texttt{allenai/Olmo-3-7B-Instruct}   & 7B  & \Checkmark & \\
                \midrule
                \multirow{2}{*}{Llama-2 \cite{touvron-llama2}}     & \texttt{Llama-2-7b-chat-hf}           & 7B  & \Checkmark & \\
                                                     & \texttt{Emollama-chat-7b}             & 7B  & \Checkmark & \Checkmark \\
                \midrule
                \multirow{2}{*}{Llama-3 \cite{grattafiori-llama3}}     & \texttt{Meta-Llama-3-8B}              & 8B  & & \\
                                                     & \texttt{Meta-Llama-3-8B-Instruct}     & 8B  & \Checkmark & \\
                \midrule
                \multirow{2}{*}{Llama-3.1 \cite{grattafiori-llama3}}          & \texttt{Meta-Llama-3.1-8B}            & 8B  & & \\
                                                     & \texttt{Meta-Llama-3.1-8B-Instruct}   & 8B  & \Checkmark & \\
                                                    \midrule
                \multirow{2}{*}{BLOOM \cite{bigscience-bloom}}      & \texttt{bloomz-7b1-mt}                & 7B  & \Checkmark & \\
                                                     & \texttt{lzw1008/Emobloom-7b}          & 7B  & \Checkmark & \Checkmark \\
                \midrule
                \multirow{2}{*}{Mistral \cite{jiang-mistral}}     & \texttt{Mistral-7B-v0.3}              & 7B  & & \\                     
                                                     & \texttt{Mistral-7B-Instruct-v0.3}     & 7B  & \Checkmark & \\
                \midrule
                Ministral\footnote{\url{https://mistral.ai/news/ministraux}}                    & \texttt{Ministral-8B-Instruct-2410}   & 8B  & \Checkmark & \\
                \midrule
                \multirow{2}{*}{Ministral 3\footnote{\url{https://mistral.ai/news/mistral-3}}} & \texttt{Ministral-3-8B-Base-2512}    & 8B  & & \\
                                                     & \texttt{Ministral-3-8B-instruct-2512} & 8B  & \Checkmark & \\
                \midrule
                \multirow{2}{*}{Qwen-2.5 \cite{qwen2.5}}    & \texttt{Qwen2.5-7B}                   & 7B  & & \\
                                                     & \texttt{Qwen2.5-7B-Instruct-1M}       & 7B  & \Checkmark & \\
                \midrule
                \multirow{2}{*}{Qwen 3 \cite{qwen3}}      & \texttt{Qwen3-4B}                     & 4B  & & \\
                                                     & \texttt{Qwen3-4B-instruct-2507}       & 4B  & \Checkmark & \\
                \midrule
                \multirow{2}{*}{DeepSeek-R1 \cite{guo-deepseek}} & \texttt{DeepSeek-R1-Distill-Llama-8B} & 8B & \Checkmark & \\
                                                     & \texttt{DeepSeek-R1-Distill-Qwen-7B}  & 7B  & \Checkmark & \\
                \bottomrule
            \end{tabular}
        }
        \caption{Summary of models evaluated throughout experiments. Emotion finetuned models are based on models released by \citet{liu-emollms}.}
        \label{tab:model-details}
    \end{table*}
}

\newcommand\enCoefStats {

    \begin{table}
        \centering
        \begin{tabular}{ccccc}
            \toprule
            Measure & Min & Max & $\mu$ & $\sigma$ \\
            \midrule
            Freq & 4.51 & 9.02 & 6.34 & 1.13 \\
            Valence & -1.00 & 0.84 & -0.43 & 0.47 \\
            Arousal & -0.96 & 0.90 & 0.11 & 0.47 \\
            Conc & 1.46 & 4.07 & 2.52 & 0.57 \\
            AoA & 3.60 & 13.41 & 8.24 & 2.76 \\
            \bottomrule
        \end{tabular}
        \caption{Summary statistics of psycholinguistic measures for the English affective states evaluated.}
        \label{tab:en-measure-stats}
    \end{table}

}

\newcommand\esCoefStats {

    \begin{table}
        \centering
        \begin{tabular}{ccccc}
            \toprule
            Measure & Min & Max & $\mu$ & $\sigma$ \\
            \midrule
            Freq & 4.21 & 8.73 & 6.19 & 1.24 \\
            Valence & -0.87 & 0.79 & -0.22 & 0.52 \\
            Arousal & -0.85 & 0.70 & 0.19 & 0.33 \\
            AoA & 3.22 & 9.98 & 6.60 & 1.65 \\
            \bottomrule
        \end{tabular}
        \caption{Summary statistics of psycholinguistic measures for the Spanish affective states evaluated.}
        \label{tab:es-measure-stats}
    \end{table}

}

\newcommand\verificationKappas {

    \begin{table}
        \centering
        \resizebox{\linewidth}{!}{
        \begin{tabular}{cccccc}
            \toprule
            \multirow{2}{*}{Model} & \multirow{2}{*}{Format} & \multicolumn{2}{c}{English} & \multicolumn{2}{c}{Spanish} \\
            \cmidrule(lr){3-4}\cmidrule(lr){5-6}
            & & $\kappa$ & $\kappa_{bin}$ & $\kappa$ & $\kappa_{bin}$ \\
            \hline
            \multirow{2}{*}{Llama 2} & Binary & 0.00 & \cellcolor{gray!80} & 0.00 & \cellcolor{gray!80} \\
            & Rating & -0.05 & -0.01 & 0.00 & 0.00 \\
            \hline
            \multirow{2}{*}{EmoLlama 2} & Binary & 0.00 & \cellcolor{gray!80} & 0.01 & \cellcolor{gray!80} \\
            & Rating & -0.06 & 0.04 & 0.10 & 0.10 \\
            \hline
            \multirow{2}{*}{OLMo} & Binary & 0.00 & \cellcolor{gray!80} & 0.00 & \cellcolor{gray!80} \\
            & Rating & 0.07 & 0.00 & -0.02 & -0.07 \\
            \hline
            \multirow{2}{*}{OLMo 2} & Binary & 0.00 & \cellcolor{gray!80} & 0.00 & \cellcolor{gray!80} \\
            & Rating & 0.13 & 0.17 & 0.08 & 0.13 \\
            \hline
            \multirow{2}{*}{OLMo 3 Base} & Binary & 0.00 & \cellcolor{gray!80} & -0.13 & \cellcolor{gray!80} \\
            & Rating & 0.22 & -0.04 & -0.07 & 0.00 \\
            \hline
            \multirow{2}{*}{OLMo 3} & Binary & -0.12 & \cellcolor{gray!80} & -0.07 & \cellcolor{gray!80} \\
            & Rating & 0.02 & 0.00 & -0.03 & 0.00 \\
            \hline
            \multirow{2}{*}{Deepseek (Llama)} & Binary & -0.07 & \cellcolor{gray!80} & 0.01 & \cellcolor{gray!80} \\
            & Rating & -0.02 & -0.08 & 0.00 & 0.00 \\
            \hline
            \multirow{2}{*}{Deepseek (Qwen)} & Binary & -0.04 & \cellcolor{gray!80} & -0.16 & \cellcolor{gray!80} \\
            & Rating & 0.05 & 0.06 & 0.03 & 0.10 \\
            \hline
            \multirow{2}{*}{Llama 3 Base} & Binary & 0.03 & \cellcolor{gray!80} & -0.08 & \cellcolor{gray!80} \\
            & Rating & 0.00 & 0.00 & 0.03 & 0.13 \\
            \hline
            \multirow{2}{*}{Llama 3} & Binary & -0.09 & \cellcolor{gray!80} & -0.07 & \cellcolor{gray!80} \\
            & Rating & 0.25 & 0.00 & 0.22 & 0.00 \\
            \hline
            \multirow{2}{*}{Llama 3.1 Base} & Binary & -0.06 & \cellcolor{gray!80} & -0.06 & \cellcolor{gray!80} \\
            & Rating & -0.04 & -0.09 & -0.02 & 0.06 \\
            \hline
            \multirow{2}{*}{Llama 3.1} & Binary & 0.04 & \cellcolor{gray!80} & -0.04 & \cellcolor{gray!80} \\
            & Rating & 0.26 & -0.09 & 0.22 & 0.42 \\
            \hline
            \multirow{2}{*}{BLOOM} & Binary & 0.01 & \cellcolor{gray!80} & -0.03 & \cellcolor{gray!80} \\
            & Rating & 0.02 & 0.09 & -0.03 & 0.38 \\
            \hline
            \multirow{2}{*}{Emo-BLOOM} & Binary & 0.01 & \cellcolor{gray!80} & 0.00 & \cellcolor{gray!80} \\
            & Rating & -0.02 & 0.17 & 0.01 & 0.21 \\
            \hline
            \multirow{2}{*}{Qwen 2.5 Base} & Binary & 0.00 & \cellcolor{gray!80} & 0.00 & \cellcolor{gray!80} \\
            & Rating & 0.04 & 0.00 & -0.06 & 0.00 \\
            \hline
            \multirow{2}{*}{Qwen 2.5} & Binary & 0.00 & \cellcolor{gray!80} & 0.00 & \cellcolor{gray!80} \\
            & Rating & 0.38 & 0.17 & 0.28 & 0.46 \\
            \hline
            \multirow{2}{*}{Qwen 3 Base} & Binary & 0.00 & \cellcolor{gray!80} & -0.04 & \cellcolor{gray!80} \\
            & Rating & 0.00 & 0.00 & 0.00 & 0.00 \\
            \hline
            \multirow{2}{*}{Qwen 3} & Binary & 0.00 & \cellcolor{gray!80} & 0.00 & \cellcolor{gray!80} \\
            & Rating & 0.13 & 0.00 & 0.06 & 0.13 \\
            \hline
            \multirow{2}{*}{Mistral Base} & Binary & -0.04 & \cellcolor{gray!80} & -0.12 & \cellcolor{gray!80} \\
            & Rating & 0.00 & 0.00 & 0.00 & 0.00 \\
            \hline
            \multirow{2}{*}{Mistral} & Binary & 0.01 & \cellcolor{gray!80} & -0.10 & \cellcolor{gray!80} \\
            & Rating & 0.23 & 0.00 & 0.18 & 0.00 \\
            \hline
            \multirow{2}{*}{Ministral} & Binary & 0.00 & \cellcolor{gray!80} & -0.08 & \cellcolor{gray!80} \\
            & Rating & 0.00 & 0.00 & 0.00 & 0.00 \\
            \hline
            \multirow{2}{*}{Ministral 3 Base} & Binary & 0.01 & \cellcolor{gray!80} & -0.12 & \cellcolor{gray!80} \\
            & Rating & 0.08 & 0.19 & 0.00 & 0.27 \\
            \hline
            \multirow{2}{*}{Ministral 3} & Binary & 0.01 & \cellcolor{gray!80} & 0.00 & \cellcolor{gray!80} \\
            & Rating & 0.19 & 0.00 & 0.22 & 0.47 \\
            \bottomrule
            \end{tabular}
        }
        \caption{Cohen's $\kappa$ scores for classification output formats in the verification task. For $\kappa_{bin}$, ratings output by the model are binarized before computing agreement.  }
        \label{tab:ver-kappas}
    \end{table}

}

\section{Introduction}

    \begin{figure}[t]
        \centering
        \includegraphics[width=0.8\linewidth]{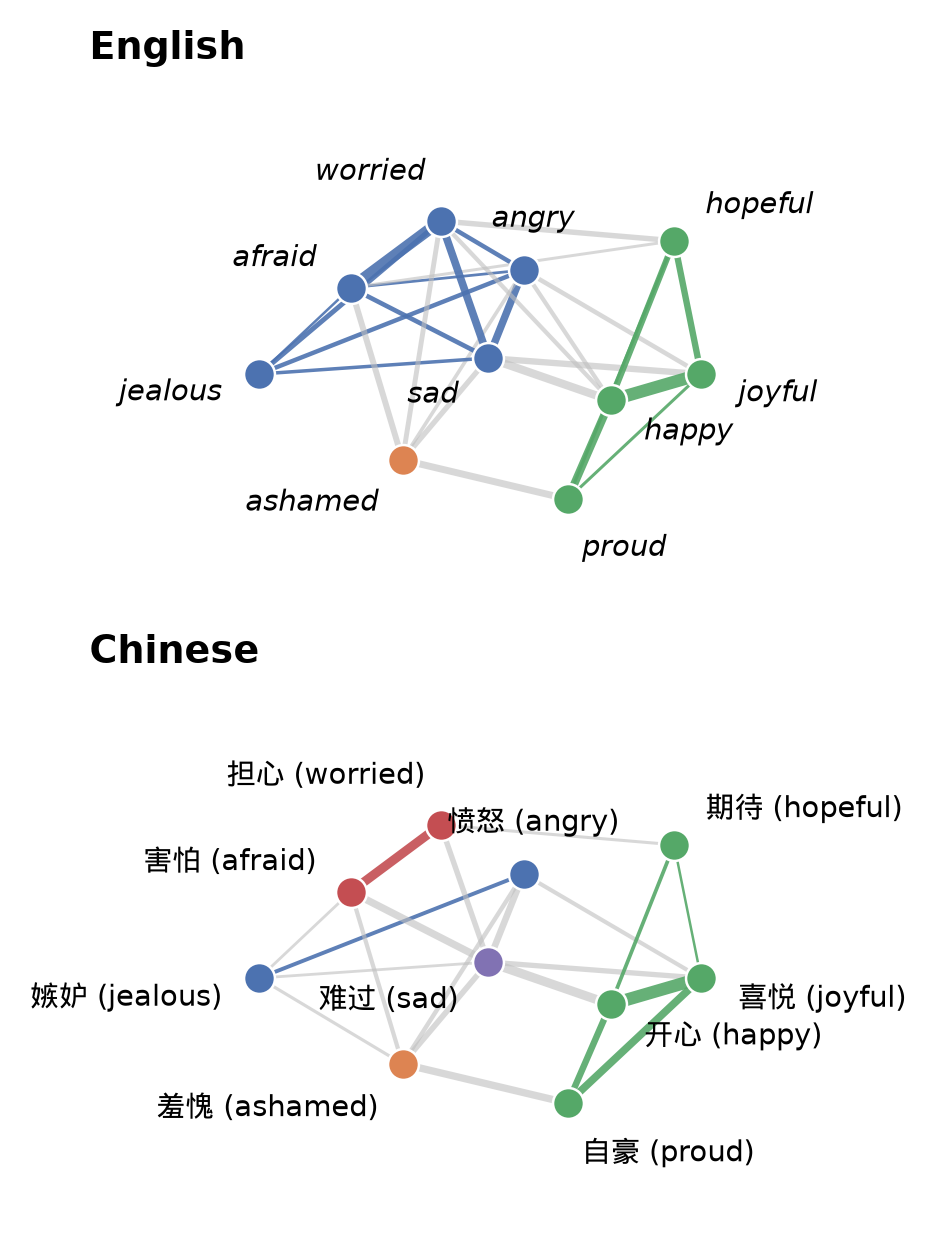}
        \caption{Subgraphs of emotion \textit{semantic relatedness networks} derived from Qwen3-8B (layer 18), where edge weights correspond to cosine similarity between contextual token representations. Colors represent communities established through Louvain community detection. The same 10 emotion concepts are shown in the same positions between networks. 
        }
        \label{fig:teaser}
    \end{figure}
 
    Given that emotion is fundamental to communication \cite{10.1093/migration/mnad018} and that  large language models (LLMs) interact with an increasingly wide and diverse group of users, it is imperative that LLMs represent and adapt to emotion 
    in culturally appropriate ways. Prior work has shown that, for example, LLMs' performance in classifying depression fails to generalize equitably across 
    language communities \cite{pereztoro2025exploring}, posing risks of mental health-related harms. More broadly, work has increasingly called for greater consideration of the close relationship between language and culture in the design of language technologies \cite{hershcovich-culture}. Not only is emotion a key proxy of this relationship \cite{adilazuarda-culture}, but emotion is itself both defined and experienced in culture-specific ways \cite{vanosch2013pride}.

    Computational linguistics work has begun to study how LLMs inherently encode emotion semantics and whether this aligns with human intuitions and expectations. This work often relies on dimensional models of emotion, where \textit{valence} and \textit{arousal} (VA) are dominant underlying factors \cite{Lindquist_Wager_Kober_Bliss-Moreau_Barrett_2012, barrett2017theory, russell-circumplex, mohammad-vad}. LLMs have been evaluated against human VA norms, among other psycholinguistic features, and have generally shown alignment with human judgments \cite{conde-psycholinguistic,martinez-mwe}, even being used to extend existing lexicons \cite{nemati-eplan}. Such evaluations, however, can suffer from the many biases and calibration concerns raised in other settings \cite{hagihara2026well}. Accordingly, other work examines emotion geometry within LLMs' activation spaces, broadly finding that VA structure aligns with human ratings of VA among basic \cite{sun-subspace} and more fine-grained \cite{shariat-mosaic} emotion concepts.

    Some literature, however, has looked to cultural variation in emotion semantics rather than universal structure. \citet{jackson-semantics} use \textit{colexification networks} (see \autoref{sec:background}), wherein multiple semantically similar concepts are expressed using a single word, to study such variation.%
    Beyond universal valence and arousal dimensions, they also surface language family-specific differences in colexification network structure.%
    Other work has highlighted specific emotion concepts that have culture-specific meaning, such as `frustration' resembling higher-power emotions (e.g., `anger') in English but lower-power emotions (e.g., `sadness') in European languages (i.e., Spanish, French, German)\cite{soriano-frustration}.~\footnote{In this context, high and low power are associated with different responses to an emotion, for example, confrontation with higher-power, withdrawal with lower-power \cite{soriano-frustration}.} 
     While computational linguistics
    has begun to examine culture-specific meanings of select emotions  (e.g., `pride'; \citealt{havaldar-multilingual}), there is a need for larger-scale systematic investigation of such variation.

    To this end, we analyze
    semantic relatedness networks of fine-grained emotion concepts from LLMs across four languages--English, Spanish, French, and Mandarin Chinese--and examine to what extent semantic relatedness patterns match those established in prior psychology and linguistics work.
   We focus on three research questions.

    \textbf{RQ1: To what extent do models encode meaningful language-specific affective state relationships \textit{within} languages?} We first conduct an exploratory analysis on monolingual emotion networks, presupposing no structure (e.g., valence/arousal dimensional models; \citealt{mohammad-vad}) to models' underlying emotion space. Inspired by \citet{jackson-semantics}'s work on colexification networks, we derive analogous \textit{semantic relatedness networks} (see English and Chinese example subnetworks for Qwen3-8B in \autoref{fig:teaser}) from models based on token-level representations of emotions, and examine whether model similarity topology mirrors established cultural and psychological intuitions. Through these networks, we surface \textbf{meaningful language-specific differences in the structure of the emotion space}. Moreover, we find that this language-dependence reflects \textit{expected} differences identified in prior psychology literature: the semantic relationships encoded between emotion concepts \textbf{largely match human conceptual similarities} reflected in \citet{jackson-semantics}'s colexification networks \textbf{as well as the specific, culturally-variable emotion concepts} in prior literature \cite{soriano-frustration,liu-pride}.  %

    Given these monolingual findings, we then consider multilingual networks, leading to our second research question.    
     \textbf{RQ2: To what extent do models encode meaningful language-specific affective state relationships \textit{between} languages?}
     We
     conduct an additional exploratory analysis on multilingual emotion networks. 
     In doing so, we find that multilingual networks show cohesive, conceptually related emotion communities. For example, 
     the translations of 
     \textit{happy} and \textit{grateful} in all four languages are all in the same community. Further supporting this, we find that translation pairs of emotion terms fall within the same communities at a rate well above chance. 

     Given that these results suggest the existence of cross-lingual affective structure, we then conduct a confirmatory probing analysis, assuming valence and arousal as the dominant underlying dimensions of emotion concepts as our third research question.
   \textbf{RQ3: To what extent do models encode meaningful language-specific \textit{valence and arousal structure}?}  Using native-speaker derived lexica, we use information theoretic probes to study how saliently native speaker intuitions of valence and arousal are encoded in models' hidden representations. 
 We find that valence is consistently more recoverable than arousal, and that recoverability aligns to some extent with expectations driven by pretraining language distributions of each model. For example, Mandarin valence and arousal are comparably decodable to French and Spanish in Qwen3, a model pretrained on a substantial amount of Mandarin Chinese data, but less decodable in Ministral3, a model with an emphasis on European languages in its training data.
 Furthermore, French and Spanish valence labels associated with translations to English are more recoverable than the labels for the original untranslated words, suggesting that English-centric valence information is more salient than that derived from native speaker judgments for those languages.

\section{Background and Related Work. }
\label{sec:background}

\paragraph{Cross-cultural Emotion Semantics and Colexification.}
Given the lack of an empirical ground truth for cross-lingual language meaning, cross-lingual and cultural variation in emotion is typically studied through an examination of relationships among emotion concepts.

For example, \citet{soriano-frustration} report the closer relationship between English `frustration' and `anger' than the translations of each in French and Spanish to concepts related to `anger' and \citet{vanosch2013pride} report that the concept of `pride' is typically interpreted more positively by Western speakers than Asian speakers in general. Similar relationships have been identified for other concepts (e.g., `anger' \cite{mikolajczuk-anger}, `shame' \cite{kollareth-shame}, `angst' \cite{wierzbicka-emotions}).

On a larger scale, \citet{jackson-semantics}  generate colexification networks for 2,439 concepts across 2,474 world languages to create a ground truth for the conceptual structure of emotion space for different language families. These networks leverage colexification, when two or more semantically similar concepts are expressed using a single word  (See \autoref{tab:colexification-examples}).

\begin{table}[h]
\centering
\small
\begin{CJK}{UTF8}{gbsn}
\begin{tabular}{@{}lll@{}}
\toprule
\textbf{Lang.} & \textbf{Word} & \textbf{Concepts expressed} \\
\midrule
\texttt{en} & \textit{draw} & pull, depict with lines \\
\texttt{fr} & \textit{femme} & woman, wife \\
\texttt{es} & \textit{pueblo} & people, village \\
\texttt{zh} & 天 \textit{tiān} & sky, heaven, day \\
\bottomrule
\end{tabular}
\end{CJK}
\caption{Colexification examples from English, French, Spanish, and Mandarin.}
\label{tab:colexification-examples}
\end{table}

In \citet{jackson-semantics}'s networks, the greater the number of languages in which a given concept pair is colexified, the stronger the edge weight between those two nodes (see reproduced network in Figure~\ref{fig:jackson-colex-network}, \autoref{app:jackson-repl}).

\paragraph{Cross-Lingual Emotion Representations.}
Past work has probed model representations for subjective concepts such as emotion, for example, \citet{havaldar-etal-2023-multilingual} and \citet{lee-etal-2025-multilingual} find that multilingual model emotion representation captures cross-lingual and cross-cultural structural variation in emotion meaning. Moreover \citet{wu2025semantic} find that models partially share representation space across languages.

\paragraph{Emotion Representation Probing}

Past work by \citealt{reichman2026emotionsartthouunderstanding} and \citealt{sun-subspace} examines emotion representation space in LLMs, and establishes that the human emotion circumplex along valence and arousal dimensions can be recovered from LLM emotion representation geometry. In this way, valence and arousal are the factors that summarize the relationships between emotions that we seek to examine. However, we consider relatedness as opposed to directly considering geometry.
\citet{havaldar-etal-2023-multilingual} begin this work by using BERT to perform limited probing experiments on multilingual conceptual representations of emotions, while our probing experiments are on a larger scale and using current models. This allows us to make broader assessments about relatedness captured in model emotion structure. 

\section{RQ1: Exploratory Monolingual Network Analysis}

    Inspired by \citet{jackson-semantics} and similar work, we conduct an exploratory analysis of the monolingual relationships encoded between emotion concepts in LLMs.
    \subsection{Data}

            \textbf{\mosaic{}. } For our analyses, we leverage \mosaic{} \cite{shariat-mosaic}, a multilingual emotion dataset of personal narratives from multiple online sources (e.g., Reddit, YouTube) with fine-grained labels derived from the authors' own description of their emotional state (i.e., \textit{``I feel X''}). The authors term these open-vocabulary self-descriptors \textit{affective states}. 

            Each sample of MOSAIC contains a \textit{context} and an \textit{affective state} label, shown in \autoref{tab:mosaic-example}.
    \begin{table}[t]
\centering
\small
\begin{tabular}{@{}lp{0.70\linewidth}@{}}
\toprule
\textbf{Field} & \textbf{Example} \\
\midrule
Affective state & \textit{uplifted} \\
Context & I feel like a dark blanket was removed from my head and I am finally able to see sunshine again. I feel \textbf{uplifted} and alive. I feel I have stronger boundaries with other people whereas before I allowed others to run all over me more easily. \\
\bottomrule
\end{tabular}
\caption{An example English \mosaic{} affective state and context pair. Each affective state is associated with multiple contexts.}
\label{tab:mosaic-example}
\end{table}

    \paragraph{Models.} We focus on three LLMs--Qwen3-8B \cite{qwen3}, Ministral-3-8B-Base-2512 \cite{liu2026ministral3}, and Gemma-3-12B-pt \cite{gemmateam2025gemma3technicalreport} given their multilinguality, the recency of their release, and that they are of typical size studied in similar research settings (e.g., \cite{xu-zhang-2026-language,deas-etal-2026-characterizing,fuxa-machacek-2026-alignatt4llm}).
    Given differences in language coverage in each of their training, we expect Ministral3 to generate more expressive and fine-grained French and Spanish representations and conversely for Qwen3 to generate more expressive and fine-grained Mandarin representations.

            \textbf{Languages.} We use the English, Spanish, Mandarin, and French subsets of \mosaic{} \cite{shariat-mosaic}.
            In addition to being widely-available, high-resource languages that have associated native speaker-derived semantic norms, this set of languages reflects both typologically similar (i.e., Spanish and French) and diverse (i.e., English and Chinese) language pairs. Moreover, it includes languages rooted in both the Western and Eastern hemisphere.
            Past literature on cultural differences in model behavior (e.g., \citealt{ananthram2025perspectivelanguageaffectscultural}) often finds systematic differences when modeling these languages. 

            \autoref{tab:data-stats} presents summary statistics of the subset of the \mosaic{} dataset we use for all experiments, including the number of states and contexts and the average length of each context in tokens for each language. To limit noise from the long tail of rare affective states with few contexts, we use the top-200 affective states and their associated contexts in all experiments, and impose a cap of 20 contexts per affective state term.

            \begin{table}[htbp]
                \centering
                \resizebox{\linewidth}{!}{
                    \begin{tabular}{cccc}
                         \toprule
                         Lang & \# States & \# Contexts & Median Context L \\
                         \midrule
                         English (\texttt{en}) & 200 & 4,000 & 60 \\
                         Spanish (\texttt{es}) & 200 & 4,000 & 90 \\
                         French (\texttt{fr})  & 200 & 3,996 & 53 \\
                         Chinese (\texttt{zh}) & 200 & 4,000 & 67 \\
                         \bottomrule
                    \end{tabular}
                }
                \caption{Summary statistics of the \mosaic{} subsets used throughout experiments. Average context length (\textit{Avg. Context L}) is calculated using the Qwen3 tokenizer. Each context is truncated to the sentence containing the emotion term, as well as the preceding and following sentence. The number of contexts per emotion term is capped at 20 for all languages. French has slightly fewer contexts than the other languages due to its last 4 emotion terms out of the 200 having 19 contexts available each.}
                \label{tab:data-stats}
            \end{table}

    \subsection{Methods} 

    Using contextual emotion representations of each affective state, we build \textit{semantic relatedness networks} from model hidden representations. We then compare the network structure for our affective state representations with reference semantic relatedness networks from \citet{jackson-semantics}. 

    \paragraph{Contextual Emotion Representations. }
   Our semantic representations capture each token's typical meaning in a variety of emotion-centric contexts.
    At each model layer, we mean-pool the residual stream hidden states of the subword tokens making up each affective state term in order to achieve a contextual representation of that term.
    We standardize these representations by subtracting the mean vector from each representation and dividing every dimension by its own standard deviation, following \citet{timkey-van-schijndel-2021-bark}, in order to correct for the effects of a small number of dimensions with outsized magnitudes dominating the representation space.
    Both the mean vector and by-dimension standard deviation are computed for each layer over all tokens in our dataset. We then average the representations over all contexts that that term appears in (up to 20).

    \paragraph{Colexification and Semantic Relatedness Networks.} Given the lack of an empirical ground truth for variation in cross-lingual semantic meaning, \citet{jackson-semantics} use colexification (\autoref{tab:colexification-examples}) to measure relatedness between 24 emotion concepts (detailed in \autoref{tab:jackson-concepts}) and study cross-lingual and cultural variation in emotion word meaning. %
    We seek to construct the counterpart for language models, leveraging the contextual affective state representations described above. We use these representations to construct monolingual \textit{semantic relatedness networks} for the 200 most common affective state terms in each of the French, English, Spanish, and Mandarin subsets of \mosaic{}, based on pairwise cosine similarity between all affective states considered within each language. 
In order to assess to what extent monolingual networks show meaningful structure, we use Louvain community detection \cite{louvain} to estimate communities of closely-related terms.

    \begin{figure*}[t]
        \centering
        \includegraphics[width=\textwidth]{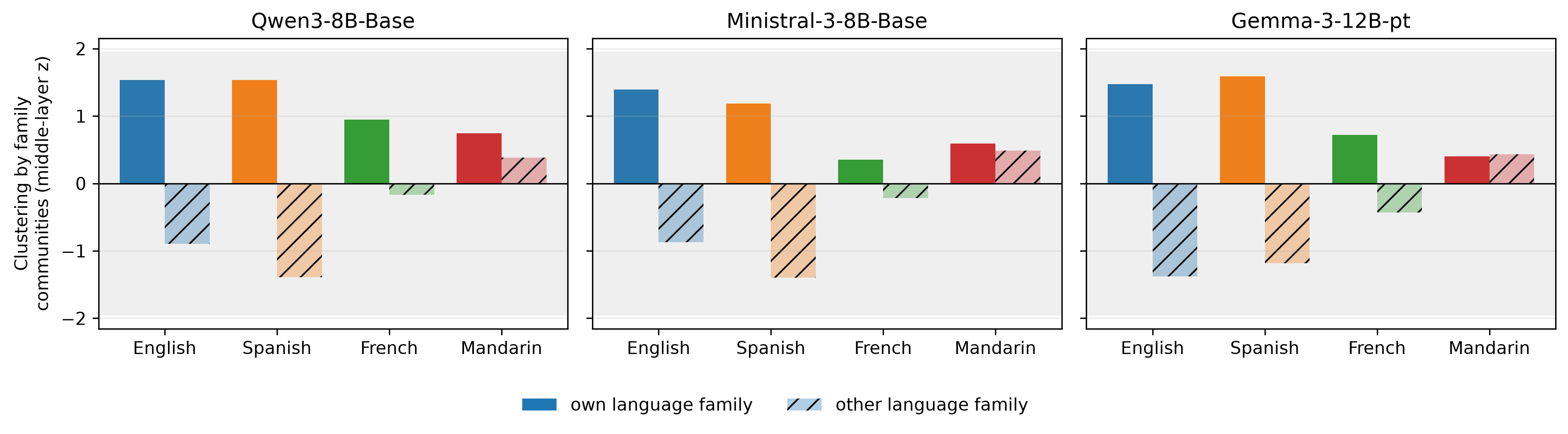}
        \caption{English, French, and Spanish affective state term similarities are statistically significantly more closely aligned with the Indo-European network relative to the Sino-Tibetan network, than Mandarin term similarities are ($p=0.011$ for Qwen3, $p=0.031$ for Ministral3, and $p=0.011$ for Gemma3).}
        \label{fig:jackson-networks}
    \end{figure*}   

    \paragraph{Jackson Language Family Community Analysis.}    
   We seek to determine whether the most similar affective state terms we identify in our cosine similarity networks mirror the communities of socio-linguistically related emotion concepts identified by \citet{jackson-semantics}. \citet{jackson-semantics} define colexification networks for the 24 emotion concepts for each global language family. Because our languages are English, French, Spanish, and Mandarin, we examine the networks for Indo-European and Sino-Tibetan languages in order to assess whether conceptual similarities within each language match those within its larger language family.
   
   For example, we examine whether the most closely-related Mandarin affective state terms are more closely aligned with the communities \citet{jackson-semantics} detect for emotion concept relatedness within the Sino-Tibetan family than they are with the Indo-European emotion concept relatedness communities.
   To do this, using the Indo-European and Sino-Tibetan networks provided by \cite{jackson-semantics}, we associate each of our top-200 terms with a given Jackson concept. For example, we map `angry', `annoyed,'`furious', `irritated,' and `pissed' of our own concepts to the Jackson concept `anger.'
 
 We then assess whether affective state terms whose Jackson concepts fall in the same community are more similar in terms of cosine similarity than those that fall in different communities. We do so by computing the mean cosine similarity between affective state terms whose associated Jackson concepts belong to the same community and subtracting the mean similarity between terms whose concepts belong to different communities. We use a null baseline built by random permutations of Jackson concept community labels to determine statistical significance.

\subsection{Results}
\paragraph{Monolingual networks show language-specific community structure.} 
After Louvain community detection, 
we see semantically meaningful emotion community formation for all languages (see \autoref{app:monolingual-networks}, \autoref{tab:monolingual_communities}). For example, at layer 18 of Qwen3, we identify positive emotion communities, including words for happiness in all four languages, and negative emotion communities, e.g. including words referring to unhappiness in all four languages. 

        \paragraph{Jackson networks suggest alignment with language-family communities}

         We use the Indo-European and Sino-Tibetan concept networks from \cite{jackson-semantics}, and find that affective state term similarities in the Indo-European languages (English, French, and Spanish) are statistically significantly more closely aligned with the Indo-European network relative to the Sino-Tibetan network, than are Mandarin term similarities ($p=0.011$ for Qwen3, $p=0.031$ for Ministral3, and $p=0.011$ for Gemma3) (\autoref{fig:jackson-networks}).

    \paragraph{Case Study: \textit{Frustration} exhibits cross-lingual variation in semantic relatedness.}

\begin{table*}[t]
\centering
\begin{CJK}{UTF8}{gbsn}
\setlength{\tabcolsep}{5pt}
\renewcommand{\arraystretch}{0.92}
\begin{subtable}[t]{0.32\textwidth}
\centering
\scriptsize
\begin{tabular}{@{}lll@{}}
\toprule
\textbf{frustrated} & \textbf{frustré} & \textbf{frustrado} \\
\midrule
\makecell[l]{annoyed \\ {\tiny\textcolor{gray}{(0.74)}}} &
\makecell[l]{vexé \\ {\tiny\textcolor{gray}{(offended; 0.54)}}} &
\makecell[l]{desanimado \\ {\tiny\textcolor{gray}{(discouraged; 0.64)}}} \\
\makecell[l]{irritated \\ {\tiny\textcolor{gray}{(0.72)}}} &
\makecell[l]{énervé \\ {\tiny\textcolor{gray}{(irritated; 0.51)}}} &
\makecell[l]{desesperado \\ {\tiny\textcolor{gray}{(desperate; 0.60)}}} \\
\makecell[l]{upset \\ {\tiny\textcolor{gray}{(0.63)}}} &
\makecell[l]{déçue \\ {\tiny\textcolor{gray}{(disappointed; 0.49)}}} &
\makecell[l]{angustiado \\ {\tiny\textcolor{gray}{(distressed; 0.58)}}} \\
\makecell[l]{angry \\ {\tiny\textcolor{gray}{(0.63)}}} &
\makecell[l]{désemparé \\ {\tiny\textcolor{gray}{(helpless; 0.49)}}} &
\makecell[l]{molesto \\ {\tiny\textcolor{gray}{(annoyed; 0.56)}}} \\
\makecell[l]{furious \\ {\tiny\textcolor{gray}{(0.62)}}} &
\makecell[l]{furieux \\ {\tiny\textcolor{gray}{(furious; 0.49)}}} &
\makecell[l]{decepcionado \\ {\tiny\textcolor{gray}{(disappointed; 0.56)}}} \\
\bottomrule
\end{tabular}
\caption{Qwen3-8B-Base, Layer 18}
\end{subtable}
\hfill
\begin{subtable}[t]{0.32\textwidth}
\centering
\scriptsize
\begin{tabular}{@{}lll@{}}
\toprule
\textbf{frustrated} & \textbf{frustré} & \textbf{frustrado} \\
\midrule
\makecell[l]{annoyed \\ {\tiny\textcolor{gray}{(0.75)}}} &
\makecell[l]{vexé \\ {\tiny\textcolor{gray}{(offended; 0.60)}}} &
\makecell[l]{angustiado \\ {\tiny\textcolor{gray}{(distressed; 0.65)}}} \\
\makecell[l]{irritated \\ {\tiny\textcolor{gray}{(0.72)}}} &
\makecell[l]{furieux \\ {\tiny\textcolor{gray}{(furious; 0.54)}}} &
\makecell[l]{desanimado \\ {\tiny\textcolor{gray}{(discouraged; 0.64)}}} \\
\makecell[l]{disappointed \\ {\tiny\textcolor{gray}{(0.69)}}} &
\makecell[l]{déçu \\ {\tiny\textcolor{gray}{(disappointed; 0.53)}}} &
\makecell[l]{molesto \\ {\tiny\textcolor{gray}{(annoyed; 0.63)}}} \\
\makecell[l]{angry \\ {\tiny\textcolor{gray}{(0.66)}}} &
\makecell[l]{déprimé \\ {\tiny\textcolor{gray}{(depressed; 0.53)}}} &
\makecell[l]{desesperado \\ {\tiny\textcolor{gray}{(desperate; 0.62)}}} \\
\makecell[l]{upset \\ {\tiny\textcolor{gray}{(0.65)}}} &
\makecell[l]{énervé \\ {\tiny\textcolor{gray}{(irritated; 0.52)}}} &
\makecell[l]{decepcionado \\ {\tiny\textcolor{gray}{(disappointed; 0.61)}}} \\
\bottomrule
\end{tabular}
\caption{Ministral-3-8B-Base, Layer 17}
\end{subtable}
\hfill
\begin{subtable}[t]{0.32\textwidth}
\centering
\scriptsize
\begin{tabular}{@{}lll@{}}
\toprule
\textbf{frustrated} & \textbf{frustré} & \textbf{frustrado} \\
\midrule
\makecell[l]{annoyed \\ {\tiny\textcolor{gray}{(0.82)}}} &
\makecell[l]{énervé \\ {\tiny\textcolor{gray}{(irritated; 0.67)}}} &
\makecell[l]{angustiado \\ {\tiny\textcolor{gray}{(distressed; 0.78)}}} \\
\makecell[l]{irritated \\ {\tiny\textcolor{gray}{(0.79)}}} &
\makecell[l]{déprimé \\ {\tiny\textcolor{gray}{(depressed; 0.66)}}} &
\makecell[l]{desanimado \\ {\tiny\textcolor{gray}{(discouraged; 0.75)}}} \\
\makecell[l]{upset \\ {\tiny\textcolor{gray}{(0.75)}}} &
\makecell[l]{vexé \\ {\tiny\textcolor{gray}{(offended; 0.66)}}} &
\makecell[l]{desesperado \\ {\tiny\textcolor{gray}{(desperate; 0.72)}}} \\
\makecell[l]{disappointed \\ {\tiny\textcolor{gray}{(0.75)}}} &
\makecell[l]{écœuré \\ {\tiny\textcolor{gray}{(disgusted; 0.63)}}} &
\makecell[l]{agobiado \\ {\tiny\textcolor{gray}{(overwhelmed; 0.70)}}} \\
\makecell[l]{angry \\ {\tiny\textcolor{gray}{(0.73)}}} &
\makecell[l]{furieux \\ {\tiny\textcolor{gray}{(furious; 0.63)}}} &
\makecell[l]{molesto \\ {\tiny\textcolor{gray}{(annoyed; 0.70)}}} \\
\bottomrule
\end{tabular}
\caption{Gemma-3-12B-pt, Layer 24}
\end{subtable}
\caption{Five nearest neighbors for frustration in English, French, and Spanish
across models, using standardized cosine similarity.
Parentheses give English glosses and cosine similarity.}
\label{tab:frustration_neighbors}

\end{CJK}
\end{table*}

        We likewise do a qualitative analysis where we observe which concepts are most closely related to `frustration' and assess whether those relationships match past work on semantic relatedness to frustration, e.g. by \citet{soriano-frustration}.

        In order to understand each language's representation space, we create proximity networks of the top-5 most semantically similar emotion concepts to `frustration.' %
        The semantically closest emotion word representations in Table \ref{tab:frustration_neighbors} demonstrate cross-lingual variation in conceptual relatedness to `frustration' for English compared to French and Spanish, reflecting the phenomena of cross-cultural variation in emotion relatedness identified by \cite{soriano-frustration}.
        Our findings match \citeauthor{soriano-frustration}'s that English `frustration' is more related to high-power anger, whereas `frustration' and `frustración' in French and Spanish are more related to lower-power words. The top-5 most similar terms to `frustrated' for English include `angry' for all models, whereas the nearest neighbors for Spanish have a much higher prevalence of lower-power words such as `desanimado' (discouraged) and `angustiado' (distressed). French likewise has a higher prevalence of lower-power related words such as `déprimé' (depressed).

    We perform a similar analysis for the concept of `pride' based on the cross-cultural semantic variation work by \citet{vanosch2013pride} (See \autoref{app:pride_case_study}).

\paragraph{Models encode appropriately language-specific affective state relationships within languages}

Taken together, our findings that monolingual networks show meaningful language-specific community structure, that Jackson networks suggest alignment with language-family communities, and that `frustration' exhibits cross-lingual variation in semantic relatedness suggest that models encode appropriately language-specific affective state relationships within languages.

\section{RQ2: Exploratory Multilingual Network Analysis}

We extend our analysis of the relationships encoded between emotion concepts in LLMs to cross-lingual networks.

\subsection{Data}

            For our analyses, we use an identical dataset to the \textbf{RQ1} experiments, derived from \mosaic{}.

    \subsection{Models}
    We use the same three LLMs--Qwen3, Ministral3, and Gemma3--as for \textbf{RQ1}.

\subsection{Methods} 

    \paragraph{Multilingual Network Analysis.}
    Using the same top 200 affective state terms for each language as for \textbf{RQ1}, we generate a cosine similarity network of all 800 terms in order to examine cross-lingual network structure. As in \textbf{RQ1}, we use Louvain community detection \cite{louvain} to define communities of closely-related terms.

    \paragraph{Regressing out language.} In order to remove the effect of within-language baseline similarity between terms, we perform ridge regression on all pairwise cosine similarities on a language indicator in order to residualize out the language-based similarity between each emotion concept representation. See appendix for community separation before and after separation (\autoref{fig:nmi}).

    \paragraph{Measuring prevalence of translation pairs in same community}
    We consider translation pairs to be terms from different languages that a given model translates into the same English word.
    
    At each layer, we compare the proportion of translation pairs in the same Louvain-detected community to a baseline probability that two randomly selected terms from the same two languages fall within the same community, taking community size into account.

\subsection{Results}

\paragraph{After residualizing out language, multilingual communities reflect cohesive affective state communities.}

Qualitatively, terms cluster into multilingual communities representing, for example, positive and negative emotion terms, and terms relating to anger or  guilt (\autoref{app:multilingual-networks}, \autoref{tab:joint_communities}).

\paragraph{Translation pairs cluster in same community after residualizing out language.}

Once language is partialled out, translation pairs cluster within the same community more often than expected by chance (\autoref{fig:translation_pairs}). For example, in the middle third layers of each model, 61--81\% of translation pairs fall in the same community, whereas before residualizing language out only 13--27\% of translation pairs fall in the same communities at these layers. For context, taking into account community size, 16--21\% would be expected by chance to fall in the same community at those layers after residualizing out language, and 5--10\% would be expected to before residualizing it out.

\paragraph{Models encode language-specific affective state relationships between languages.}

After partialling out language, multilingual communities reflect cohesive affective state communities (\autoref{app:multilingual-networks}, \autoref{tab:joint_communities}), and translation pairs cluster in the same community at a rate well above chance, suggesting that models encode language-specific affective state relationships between languages.

\section{RQ3) Confirmatory Valence-Arousal Analysis}

    In \textbf{RQ1} and \textbf{RQ2} above, we use an inductive approach to broadly explore similarities and differences in emotion representations across languages. To complement this analysis, we now conduct a confirmatory analysis based on language-specific valence and arousal dimensions underlying affective states to empirically assess structural differences. As in \textbf{RQ2}, we consider cross-lingual relationships, but we now turn to analysis of overarching structure, rather than examining relationships between individual pairs of terms. In order to assess whether models encode valence and arousal in culturally-conditioned ways, we train a probe on activations representing contexts to predict valence and arousal associated with affective state. 

\subsection{Methods}

\begin{figure*}[t]
        \centering
        \includegraphics[width=0.8\linewidth, trim = 1cm 0cm 0cm 0cm]{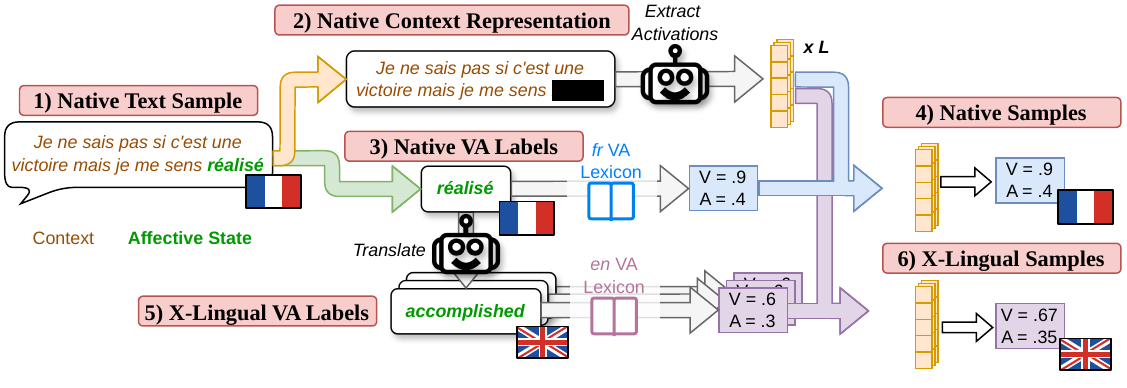}
        \caption{Illustration of the intra- and cross-lingual probing design.}
        \label{fig:xlingual-probe-diagram}
    \end{figure*}

    Our overall approach to probing valence and arousal both intra- and cross-lingually is illustrated in \autoref{fig:xlingual-probe-diagram}. We use the same models and underlying data for these experiments.

    \paragraph{Context Representations. }
    Given a text sample in a particular language from \mosaic{} (\textbf{1} in \autoref{fig:xlingual-probe-diagram}), we extract the residual stream activation of the last subword token of each affective state term at each layer (\textbf{2}).
    We standardize these representations in the same way as we standardized the representations from \textbf{RQ1} and \textbf{RQ2}, following \citet{timkey-van-schijndel-2021-bark}. The representation from each context is used as a sample.

     \paragraph{Psychometric Lexicons. } We use native speaker-derived English (NRC-VAD v2.1; \citealt{mohammad-vad-v2}), Spanish \cite{gonzalez-vad}, French (MEmoLon; \citealt{buechel-memolon}), and Mandarin Chinese (CVAW; \citealt{lee2022chinese,yu2016building}) VAD lexicons--dictionaries mapping terms to average valence, arousal norms--in order to capture the emotion-related intuitions of native speakers. Using labels derived from native speakers, as opposed to translations, better captures the cultural context of language terms. 

     Valence, meaning positivity of the emotion, and arousal, meaning intensity of emotion, give us a set of metrics by which to consider emotion concepts. These annotations serve as the labels (\textbf{3}) for our cross-lingual probing analyses of emotion concept representations.
            
    Summary statistics and details on the psychometric lexica used are included in \autoref{app:lexicon-dets}. Generally, these lexica cover 80\% (Spanish) to 100\% (French) of the affective states we consider.

    \paragraph{MDL probing.}
    With enough training, linear probing can conflate memorization of information with trivially recoverable (i.e., explicitly encoded) information. 
    In addition, cross-lingual comparison is made challenging by inherent differences in syntax, tokenization, and data coverage among others. In part to address such issues, \citet{bugliarello-etal-2020-easier} use an information-theoretic metric to measure the difficulty of translating between different language pairs. In this vein, we employ MDL probing \cite{voita-titov-2020-information}, a method grounded in information theory that simultaneously mitigates the effects of probe memorization and better enables cross-lingual comparison. Since MDL probing reflects the amount of data needed for probe accuracy, it can capture how cross-lingual differences in language syntax, tokenization, and data coverage affect recoverability of valence and arousal information. Details of our implementation can be found in \autoref{app:mdl-probing}.

    \begin{figure*}[t]
        \centering
        \includegraphics[width=0.85\linewidth]{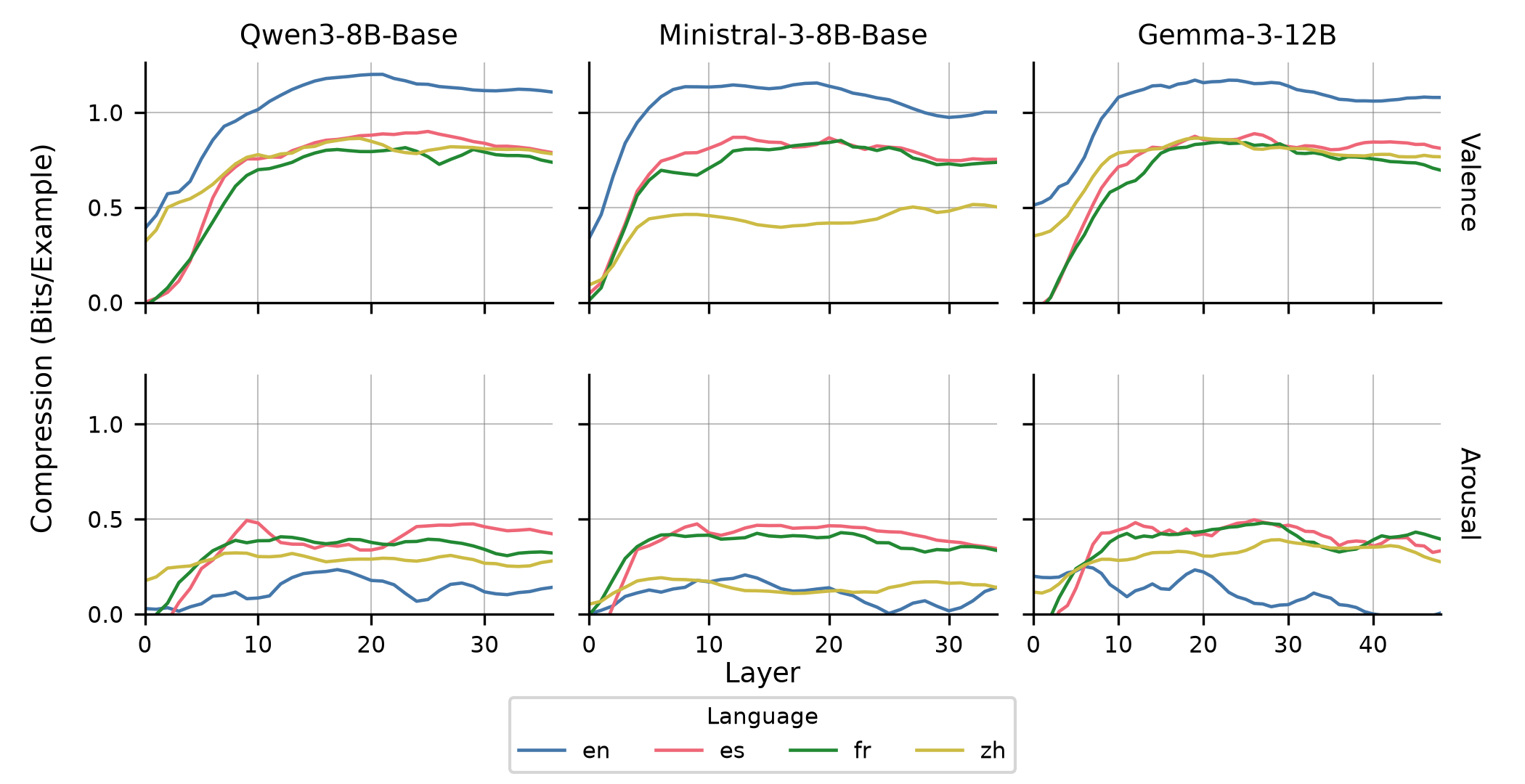}
        \caption{MDL compression for each model and language. }
        \label{fig:compression}
    \end{figure*}

    \paragraph{Cross-Lingual Probing. }
    The process described trains probes on \textit{native} samples (\textbf{4})--pairs of context representations and VA scores of the associated affective state all in a specific language. In order to additionally assess to what extent valence and arousal information is shared cross-lingually, we train probes on each languages' representations to recover English valence and arousal labels. In \autoref{fig:xlingual-probe-diagram}, we take the top candidate English translations from a given model, calculate the average VA scores based on those translations using the English lexicon (\textbf{5}), and form cross-lingual probe training samples (\textbf{6}) using these scores and the original representations based on untranslated contexts.
    
    To summarize how models encode emotion representations cross-lingually, we compare language-specific and translation-based probes. We calculate \textit{translation recovery}--the percentage of compression recovered after the aforementioned translation of labels to English. In other words, we assess the extent to which using the models' English translations improves compression of valence and arousal information. Specifically, if $\mathcal{C}$ is the compression in bits/example, we calculate:
        \begin{align}
            \text{\% Recovered } &= \frac{C_{xx\rightarrow en} - C_{xx}}{C_{en} - C_{xx}}
        \end{align}
    
\section{Results}

\paragraph{Valence is more recoverable than arousal.}
The amount of compression of valence labels in representations is consistently higher than the amount of compression of arousal labels for all models (\autoref{fig:compression}), suggesting that arousal is consistently more weakly encoded.

\paragraph{Mandarin valence recoverability is weaker on Ministral3.}
For Ministral3, Mandarin valence compression is lower than valence compression for any other language (\autoref{fig:compression}). On the other hand, Mandarin valence compression is on a similar level with French and Spanish for the other two models. We hypothesize this may be due to the relative weakness of Ministral3 on Mandarin due to its limited Mandarin data coverage.

\begin{figure}[tbp]
        \centering
        \includegraphics[width=0.85\linewidth]{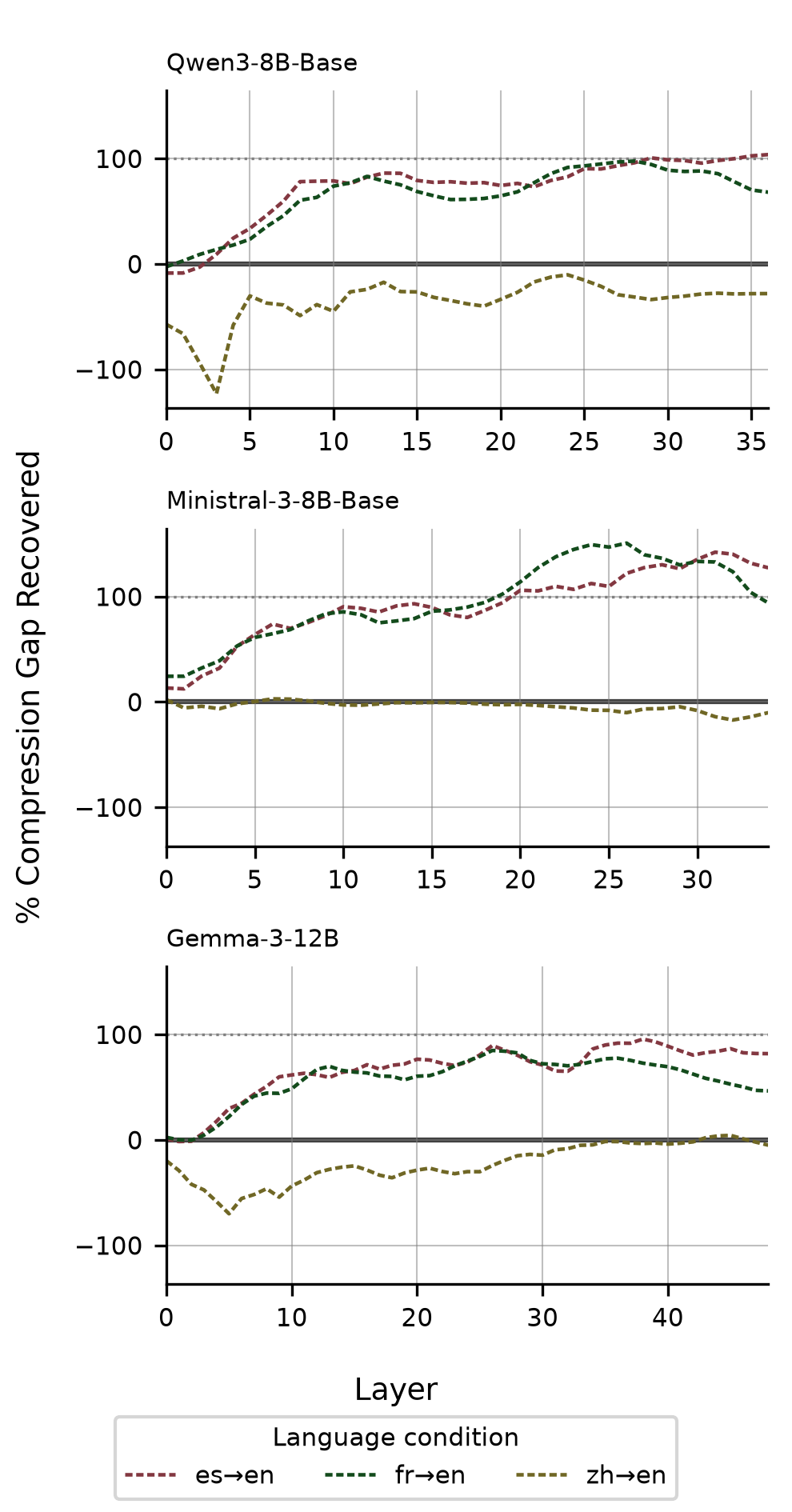}
        \caption{\% MDL compression recovered when translating lexicons from language \textit{X} into English (\texttt{xx}$\rightarrow$\texttt{en}). }
        \label{fig:recovery}
    \end{figure}

\paragraph{English-derived labels can improve recoverability.}
For French and Spanish, valence scores associated with translations to English achieve higher compression than labels associated with the original untranslated words (\autoref{fig:recovery}). This is consistent with past work, for example by \citet{lu-etal-2025-paths}, about the English-centric view of multilingual models.
Notably, however, this pattern does not hold for Mandarin. For Ministral3, the recovered compression gap under English translation is approximately 0 and for Gemma3 it is negative for the early and middle layers, while for Qwen3 translation to English generally worsens compression.
This suggests that the relative recoverability of affective information is both language-dependent and model-dependent.

\paragraph{Models encode language-specific valence and arousal structure.}
Our finding that the recoverability of affective information is language-dependent supports the idea that models encode language-specific valence and arousal structure.

\section{Discussion and Conclusion}

We study the degree to which mono- and multilingual emotion representation networks capture meaningful conceptual relationships between emotions, and to what extent valence and arousal act as principal axes of summarizing covariance and similarity between emotion concepts. From an exploratory perspective, we set out to answer \textbf{RQ1: To what extent do models encode appropriately language-specific affective state relationships within languages?}

We find that monolingual networks show language-specific community structure, and that language networks are aligned to some extent with language-family communities established by \cite{jackson-semantics}. Moreover, we find that cross-lingual variation in conceptual relatedness to both `frustration' and `pride' reflects the phenomena of cross-cultural variation in emotion relatedness identified by \cite{soriano-frustration} and \cite{vanosch2013pride}. 
We then examine cross-lingual network structure in our second research question. \textbf{RQ2: To what extent do models encode meaningful language-specific affective state relationships \textit{between} languages?} 

We find that after residualizing out the baseline similarity of words in the same language, multilingual communities reflect meaningful affective state communities \autoref{tab:joint_communities}, and translation pairs cluster in the same community at a rate above chance, suggesting that models do encode language-specific affective state relationships between languages.
Whereas \textbf{RQ1} and \textbf{RQ2} use an inductive approach to explore similarities and differences in emotion representations across languages, we seek to  analyze overarching structure in our third research question, a confirmatory valence and arousal probing analysis. \textbf{RQ3: To what extent do models encode appropriately language-specific valence and arousal structure?}

Our probing results highlight valence as the principal axis along which both shared network structure and cross-lingual variation in networks are captured. Moreover, our finding that the relative recoverability of affective information is language-dependent suggests that models encode language-specific valence and arousal structure. Our deductive methodology also reveals model-specific performance matching our expectations given differences in language coverage in each of Qwen3, Ministral3, and Gemma3, and reinforces past findings (e.g. \citet{lu-etal-2025-paths}) about the English-centric view of multilingual models.

\section*{Limitations}

Given that our analysis centers around English, French, Spanish, and Mandarin representations, future work should consider emotion semantic representations in other languages. In particular, future work should perform network topology analysis for emotion terms in lower-resource languages.
Moreover, future work should examine networks between representations generated by models from other families and of different sizes than those we employ in this work. With this, the actual pretraining distributions of the models we use are not publicly accessible, meaning that we infer their strongest languages. In addition, given our use of a non-native-speaker annotated valence-arousal lexicon for French, we do not have full native-speaker annotated coverage across all languages for these lexica.

\bibliography{custom}

\begin{thebibliography}{44}
\providecommand{\natexlab}[1]{#1}

\bibitem[{Adilazuarda et~al.(2024)Adilazuarda, Mukherjee, Lavania, Singh, Aji, O{'}Neill, Modi, and Choudhury}]{adilazuarda-culture}
Muhammad~Farid Adilazuarda, Sagnik Mukherjee, Pradhyumna Lavania, Siddhant~Shivdutt Singh, Alham~Fikri Aji, Jacki O{'}Neill, Ashutosh Modi, and Monojit Choudhury. 2024.
\newblock \href {https://doi.org/10.18653/v1/2024.emnlp-main.882} {Towards measuring and modeling ``culture'' in {LLM}s: A survey}.
\newblock In \emph{Proceedings of the 2024 Conference on Empirical Methods in Natural Language Processing}, pages 15763--15784, Miami, Florida, USA. Association for Computational Linguistics.

\bibitem[{Ananthram et~al.(2025)Ananthram, Stengel-Eskin, Bansal, and McKeown}]{ananthram2025perspectivelanguageaffectscultural}
Amith Ananthram, Elias Stengel-Eskin, Mohit Bansal, and Kathleen McKeown. 2025.
\newblock \href {https://arxiv.org/abs/2406.11665} {See it from my perspective: How language affects cultural bias in image understanding}.
\newblock \emph{Preprint}, arXiv:2406.11665.

\bibitem[{Anonymized(2026)}]{shariat-mosaic}
Anonymized. 2026.
\newblock {MOSAIC-EMO}: Multilingual large language models’ understanding of nuanced affect across cultures.
\newblock In \emph{Under Review}.

\bibitem[{Barrett(2017)}]{barrett2017theory}
Lisa~Feldman Barrett. 2017.
\newblock \href {https://doi.org/10.1093/scan/nsx060} {The theory of constructed emotion: An active inference account of interoception and categorization}.
\newblock \emph{Social Cognitive and Affective Neuroscience}, 12(11):1833--1833.

\bibitem[{Blondel et~al.(2008)Blondel, Guillaume, Lambiotte, and Lefebvre}]{louvain}
Vincent Blondel, Jean-Loup Guillaume, Renaud Lambiotte, and Etienne Lefebvre. 2008.
\newblock \href {https://doi.org/10.1088/1742-5468/2008/10/P10008} {Fast unfolding of communities in large networks}.
\newblock \emph{Journal of Statistical Mechanics: Theory and Experiment}, 2008:P10008.

\bibitem[{Buechel et~al.(2020)Buechel, R{\"u}cker, and Hahn}]{buechel-memolon}
Sven Buechel, Susanna R{\"u}cker, and Udo Hahn. 2020.
\newblock \href {https://doi.org/10.18653/v1/2020.acl-main.112} {Learning and evaluating emotion lexicons for 91 languages}.
\newblock In \emph{Proceedings of the 58th Annual Meeting of the Association for Computational Linguistics}, pages 1202--1217, Online. Association for Computational Linguistics.

\bibitem[{Bugliarello et~al.(2020)Bugliarello, Mielke, Anastasopoulos, Cotterell, and Okazaki}]{bugliarello-etal-2020-easier}
Emanuele Bugliarello, Sabrina~J. Mielke, Antonios Anastasopoulos, Ryan Cotterell, and Naoaki Okazaki. 2020.
\newblock \href {https://doi.org/10.18653/v1/2020.acl-main.149} {It{'}s easier to translate out of {E}nglish than into it: {M}easuring neural translation difficulty by cross-mutual information}.
\newblock In \emph{Proceedings of the 58th Annual Meeting of the Association for Computational Linguistics}, pages 1640--1649, Online. Association for Computational Linguistics.

\bibitem[{Conde et~al.(2025)Conde, Saiz, Grandury, Reviriego, Mart{\'i}nez, and Brysbaert}]{conde-psycholinguistic}
Javier Conde, Miguel~Gonz{\'a}lez Saiz, Mar{\'i}a Grandury, Pedro Reviriego, Gonzalo Mart{\'i}nez, and Marc Brysbaert. 2025.
\newblock \href {https://aclanthology.org/2025.gem-1.2/} {Psycholinguistic word features: a new approach for the evaluation of {LLM}s alignment with humans}.
\newblock In \emph{Proceedings of the Fourth Workshop on Generation, Evaluation and Metrics (GEM{\texttwosuperior})}, pages 8--17, Vienna, Austria and virtual meeting. Association for Computational Linguistics.

\bibitem[{Deas et~al.(2026)Deas, Mejia, Yang, and McKeown}]{deas-etal-2026-characterizing}
Nicholas Deas, Ivan Ernesto~Perez Mejia, Ellie Yang, and Kathleen McKeown. 2026.
\newblock \href {https://doi.org/10.18653/v1/2026.acl-long.2188} {Characterizing and evaluating working emotion vocabularies in multilingual large language models}.
\newblock In \emph{Proceedings of the 64th Annual Meeting of the {A}ssociation for {C}omputational {L}inguistics (Volume 1: Long Papers)}, pages 47272--47294, San Diego, California, United States. Association for Computational Linguistics.

\bibitem[{Dennison(2024)}]{10.1093/migration/mnad018}
James Dennison. 2024.
\newblock \href {https://doi.org/10.1093/migration/mnad018} {Emotions: functions and significance for attitudes, behaviour, and communication}.
\newblock \emph{Migration Studies}, 12(1):1--20.

\bibitem[{Fuxa and Mach{\'a}{\v{c}}ek(2026)}]{fuxa-machacek-2026-alignatt4llm}
Quentin Fuxa and Dominik Mach{\'a}{\v{c}}ek. 2026.
\newblock \href {https://doi.org/10.18653/v1/2026.iwslt-1.32} {{A}lign{A}tt4{LLM}: Fast {A}lign{A}tt for decoder-only {LLM}s at {IWSLT} 2026 simultaneous speech translation task}.
\newblock In \emph{Proceedings of the 23rd International Conference on Spoken Language Translation ({IWSLT} 2026)}, pages 284--295, San Diego, USA (in-person and online). Association for Computational Linguistics.

\bibitem[{Hagihara and Miyazawa(2026)}]{hagihara2026well}
Hiromichi Hagihara and Kazuki Miyazawa. 2026.
\newblock \href {https://doi.org/10.3758/s13428-025-02938-2} {How well do large language models mirror human cognition of word concepts? a comparison of psychological ratings for early-acquired english words}.
\newblock \emph{Behavior Research Methods}, 58(2):58.

\bibitem[{Havaldar et~al.(2023{\natexlab{a}})Havaldar, Rai, Singhal, Liu, Guntuku, and Ungar}]{havaldar-etal-2023-multilingual}
Shreya Havaldar, Sunny Rai, Bhumika Singhal, Langchen Liu, Sharath~Chandra Guntuku, and Lyle Ungar. 2023{\natexlab{a}}.
\newblock \href {https://doi.org/10.18653/v1/2023.wassa-1.19} {Multilingual language models are not multicultural: A case study in emotion}.
\newblock In \emph{Proceedings of the 13th Workshop on Computational Approaches to Subjectivity, Sentiment, {\&} Social Media Analysis}, pages 202--214, Toronto, Canada. Association for Computational Linguistics.

\bibitem[{Havaldar et~al.(2023{\natexlab{b}})Havaldar, Singhal, Rai, Liu, Guntuku, and Ungar}]{havaldar-multilingual}
Shreya Havaldar, Bhumika Singhal, Sunny Rai, Langchen Liu, Sharath~Chandra Guntuku, and Lyle Ungar. 2023{\natexlab{b}}.
\newblock \href {https://doi.org/10.18653/v1/2023.wassa-1.19} {Multilingual language models are not multicultural: A case study in emotion}.
\newblock In \emph{Proceedings of the 13th Workshop on Computational Approaches to Subjectivity, Sentiment, {\&} Social Media Analysis}, pages 202--214, Toronto, Canada. Association for Computational Linguistics.

\bibitem[{Hershcovich et~al.(2022)Hershcovich, Frank, Lent, de~Lhoneux, Abdou, Brandl, Bugliarello, Cabello~Piqueras, Chalkidis, Cui, Fierro, Margatina, Rust, and S{\o}gaard}]{hershcovich-culture}
Daniel Hershcovich, Stella Frank, Heather Lent, Miryam de~Lhoneux, Mostafa Abdou, Stephanie Brandl, Emanuele Bugliarello, Laura Cabello~Piqueras, Ilias Chalkidis, Ruixiang Cui, Constanza Fierro, Katerina Margatina, Phillip Rust, and Anders S{\o}gaard. 2022.
\newblock \href {https://doi.org/10.18653/v1/2022.acl-long.482} {Challenges and strategies in cross-cultural {NLP}}.
\newblock In \emph{Proceedings of the 60th Annual Meeting of the Association for Computational Linguistics (Volume 1: Long Papers)}, pages 6997--7013, Dublin, Ireland. Association for Computational Linguistics.

\bibitem[{Jackson et~al.(2019)Jackson, Watts, Henry, List, Forkel, Mucha, Greenhill, Gray, and Lindquist}]{jackson-semantics}
Joshua~Conrad Jackson, Joseph Watts, Teague~R. Henry, Johann-Mattis List, Robert Forkel, Peter~J. Mucha, Simon~J. Greenhill, Russell~D. Gray, and Kristen~A. Lindquist. 2019.
\newblock \href {https://doi.org/10.1126/science.aaw8160} {Emotion semantics show both cultural variation and universal structure}.
\newblock \emph{Science}, 366(6472):1517--1522.

\bibitem[{Kollareth et~al.(2018)Kollareth, Fernandez-Dols, and Russell}]{kollareth-shame}
Dolichan Kollareth, Jose-Miguel Fernandez-Dols, and James~A. Russell. 2018.
\newblock \href {https://doi.org/10.1163/15685373-12340031} {Shame as a culture-specific emotion concept}.
\newblock \emph{Journal of Cognition and Culture}, 18(3-4):274–292.

\bibitem[{Lee et~al.(2022)Lee, Li, and Yu}]{lee2022chinese}
Lung-Hao Lee, Jian-Hong Li, and Liang-Chih Yu. 2022.
\newblock Chinese emobank: Building valence-arousal resources for dimensional sentiment analysis.
\newblock \emph{ACM Transactions on Asian and Low-Resource Language Information Processing}, 21(4):1--18.

\bibitem[{Lee et~al.(2025)Lee, Sim, Kim, and Kim}]{lee-etal-2025-multilingual}
Woojin Lee, Yujin Sim, Hongjin Kim, and Harksoo Kim. 2025.
\newblock \href {https://doi.org/10.18653/v1/2025.ijcnlp-long.44} {Multilingual, not multicultural: Uncovering the cultural empathy gap in {LLM}s through a comparative empathetic dialogue benchmark}.
\newblock In \emph{Proceedings of the 14th International Joint Conference on Natural Language Processing and the 4th Conference of the Asia-Pacific Chapter of the Association for Computational Linguistics}, pages 791--809, Mumbai, India. The Asian Federation of Natural Language Processing and The Association for Computational Linguistics.

\bibitem[{Lindquist et~al.(2012)Lindquist, Wager, Kober, Bliss-Moreau, and Barrett}]{Lindquist_Wager_Kober_Bliss-Moreau_Barrett_2012}
Kristen~A. Lindquist, Tor~D. Wager, Hedy Kober, Eliza Bliss-Moreau, and Lisa~Feldman Barrett. 2012.
\newblock \href {https://doi.org/10.1017/S0140525X11000446} {The brain basis of emotion: A meta-analytic review}.
\newblock \emph{Behavioral and Brain Sciences}, 35(3):121–143.

\bibitem[{Liu et~al.(2026)Liu, Khandelwal, Subramanian, Jouault, Rastogi, Sadé, Jeffares, Jiang, Cahill, Gavaudan, Sablayrolles, Héliou, You, Ehrenberg, Lo, Eliseev, Calvi, Sooriyarachchi, Bout, Rozière, Monicault, Lanfranchi, Barreau, Courtot, Grattarola, Dabert, de~las Casas, Chane-Sane, Ahmed, Berrada, Ecrepont, Guinet, Novikov, Kunsch, Lample, Martin, Gupta, Ludziejewski, Rute, Studnia, Amar, Delas, Roberts, Yadav, Chandu, Jain, Aitchison, Fainsin, Blier, Zhao, Martin, Saulnier, Gao, Buyl, Jennings, Pellat, Prins, Poirée, Guillaumin, Dinot, Futeral, Darrin, Augustin, Chiquier, Schimpf, Grinsztajn, Gupta, Raghuraman, Bousquet, Duchenne, Wang, von Platen, Jacob, Wambergue, Kurylowicz, Muddireddy, Chagniot, Stock, Agrawal, Torroba, Sauvestre, Soletskyi, Menneer, Vaze, Barry, Gandhi, Waghjale, Gandhi, Ghosh, Mishra, Aithal, Antoniak, Scao, Cachet, Sorg, Lavril, Saada, Chabal, Foubert, Robert, Wang, Lawson, Bewley, Bewley, Edwards, Jamil, Tomasini, Nemychnikova, Phung, Maladière, Richard, Bouaziz, Li,
  Marshall, Li, Yang, Ouahidi, Wang, Tang, and Ramzi}]{liu2026ministral3}
Alexander~H. Liu, Kartik Khandelwal, Sandeep Subramanian, Victor Jouault, Abhinav Rastogi, Adrien Sadé, Alan Jeffares, Albert Jiang, Alexandre Cahill, Alexandre Gavaudan, Alexandre Sablayrolles, Amélie Héliou, Amos You, Andy Ehrenberg, Andy Lo, Anton Eliseev, Antonia Calvi, Avinash Sooriyarachchi, Baptiste Bout, Baptiste Rozière, Baudouin~De Monicault, Clémence Lanfranchi, Corentin Barreau, Cyprien Courtot, Daniele Grattarola, Darius Dabert, Diego de~las Casas, Elliot Chane-Sane, Faruk Ahmed, Gabrielle Berrada, Gaëtan Ecrepont, Gauthier Guinet, Georgii Novikov, Guillaume Kunsch, Guillaume Lample, Guillaume Martin, Gunshi Gupta, Jan Ludziejewski, Jason Rute, Joachim Studnia, Jonas Amar, Joséphine Delas, Josselin~Somerville Roberts, Karmesh Yadav, Khyathi Chandu, Kush Jain, Laurence Aitchison, Laurent Fainsin, Léonard Blier, Lingxiao Zhao, Louis Martin, Lucile Saulnier, Luyu Gao, Maarten Buyl, Margaret Jennings, Marie Pellat, Mark Prins, Mathieu Poirée, Mathilde Guillaumin, Matthieu Dinot, Matthieu
  Futeral, Maxime Darrin, Maximilian Augustin, Mia Chiquier, Michel Schimpf, Nathan Grinsztajn, Neha Gupta, Nikhil Raghuraman, Olivier Bousquet, Olivier Duchenne, Patricia Wang, Patrick von Platen, Paul Jacob, Paul Wambergue, Paula Kurylowicz, Pavankumar~Reddy Muddireddy, Philomène Chagniot, Pierre Stock, Pravesh Agrawal, Quentin Torroba, Romain Sauvestre, Roman Soletskyi, Rupert Menneer, Sagar Vaze, Samuel Barry, Sanchit Gandhi, Siddhant Waghjale, Siddharth Gandhi, Soham Ghosh, Srijan Mishra, Sumukh Aithal, Szymon Antoniak, Teven~Le Scao, Théo Cachet, Theo~Simon Sorg, Thibaut Lavril, Thiziri~Nait Saada, Thomas Chabal, Thomas Foubert, Thomas Robert, Thomas Wang, Tim Lawson, Tom Bewley, Tom Bewley, Tom Edwards, Umar Jamil, Umberto Tomasini, Valeriia Nemychnikova, Van Phung, Vincent Maladière, Virgile Richard, Wassim Bouaziz, Wen-Ding Li, William Marshall, Xinghui Li, Xinyu Yang, Yassine~El Ouahidi, Yihan Wang, Yunhao Tang, and Zaccharie Ramzi. 2026.
\newblock \href {https://arxiv.org/abs/2601.08584} {Ministral 3}.
\newblock \emph{Preprint}, arXiv:2601.08584.

\bibitem[{Liu et~al.(2021)Liu, Li, Chen, Wu, Yuan, and Yu}]{liu-pride}
Conghui Liu, Jing Li, Chuansheng Chen, Hanlin Wu, Li~Yuan, and Guoliang Yu. 2021.
\newblock \href {https://doi.org/10.3389/fpsyg.2021.513779} {Individual pride and collective pride: Differences between chinese and american corpora}.
\newblock \emph{Frontiers in Psychology}, Volume 12 - 2021.

\bibitem[{Lu et~al.(2025)Lu, Zhang, Eickhoff, and Pavlick}]{lu-etal-2025-paths}
Meng Lu, Ruochen Zhang, Carsten Eickhoff, and Ellie Pavlick. 2025.
\newblock \href {https://doi.org/10.18653/v1/2025.emnlp-main.762} {Paths not taken: Understanding and mending the multilingual factual recall pipeline}.
\newblock In \emph{Proceedings of the 2025 Conference on Empirical Methods in Natural Language Processing}, pages 15066--15096, Suzhou, China. Association for Computational Linguistics.

\bibitem[{Martínez et~al.(2024)Martínez, Molero, González, Conde, Brysbaert, and Reviriego}]{martinez-mwe}
Gonzalo Martínez, Juan~Diego Molero, Sandra González, Javier Conde, Marc Brysbaert, and Pedro Reviriego. 2024.
\newblock \href {https://doi.org/10.3758/s13428-024-02515-z} {Using large language models to estimate features of multi-word expressions: Concreteness, valence, arousal}.
\newblock \emph{Behavior Research Methods}, 57(1).

\bibitem[{Mikolajczuk(2000)}]{mikolajczuk-anger}
Agnieszka Mikolajczuk. 2000.
\newblock \href {https://theses.gla.ac.uk/73405/} {The conceptualisation of anger in {English} and {Polish}: A comparative study}.
\newblock {MPhil(R)} thesis, University of Glasgow, Glasgow.
\newblock Enlighten Theses, eprint 73405; ProQuest Dissertations \& Theses AAI10647946.

\bibitem[{Mohammad(2018)}]{mohammad-vad}
Saif~M. Mohammad. 2018.
\newblock Obtaining reliable human ratings of valence, arousal, and dominance for 20,000 english words.
\newblock In \emph{Proceedings of The Annual Conference of the Association for Computational Linguistics (ACL)}, Melbourne, Australia.

\bibitem[{Mohammad(2025)}]{mohammad-vad-v2}
Saif~M. Mohammad. 2025.
\newblock \href {https://arxiv.org/abs/2503.23547} {Nrc vad lexicon v2: Norms for valence, arousal, and dominance for over 55k english terms}.
\newblock \emph{Preprint}, arXiv:2503.23547.

\bibitem[{Nemati et~al.(2026)Nemati, Westbury, Rostami, and Alavi}]{nemati-eplan}
Fatemeh Nemati, Chris Westbury, Habib Rostami, and Fatemeh Alavi. 2026.
\newblock \href {https://doi.org/10.3758/s13428-026-02963-9} {Extrapolated persian lexical affect norms (e-plan) from best–worst judgments of valence, arousal, dominance, and concreteness}.
\newblock \emph{Behavior Research Methods}, 58(5).

\bibitem[{P{\'e}rez-Toro et~al.(2025)P{\'e}rez-Toro, Dineley, Iniesta, Zhang, Matcham, Siddi, Lamers, Haro, Penninx, Folarin, Arias-Vergara, Orozco-Arroyave, N{\"o}th, Maier, Wykes, Vairavan, Dobson, Narayan, Hotopf, and Cummins}]{pereztoro2025exploring}
Paula~Andrea P{\'e}rez-Toro, Judith Dineley, Raquel Iniesta, Yuezhou Zhang, Faith Matcham, Sara Siddi, Femke Lamers, Josep~Maria Haro, Brenda W. J.~H. Penninx, Amos~A. Folarin, Tom{\'a}s Arias-Vergara, Juan~Rafael Orozco-Arroyave, Elmar N{\"o}th, Andreas Maier, Til Wykes, Srinivasan Vairavan, Richard Dobson, Vaibhav~A. Narayan, Matthew Hotopf, and Nicholas Cummins. 2025.
\newblock \href {https://doi.org/10.1038/s41598-025-19980-x} {Exploring biases related to the use of large language models in a multilingual depression corpus}.
\newblock \emph{Scientific Reports}, 15(1):36197.

\bibitem[{Reichman et~al.(2026)Reichman, Avsian, and Heck}]{reichman2026emotionsartthouunderstanding}
Benjamin Reichman, Adar Avsian, and Larry Heck. 2026.
\newblock \href {https://arxiv.org/abs/2510.22042} {Emotions where art thou: Understanding and characterizing the emotional latent space of large language models}.
\newblock \emph{Preprint}, arXiv:2510.22042.

\bibitem[{Russell(1980)}]{russell-circumplex}
James~A. Russell. 1980.
\newblock \href {https://doi.org/10.1037/h0077714} {A circumplex model of affect.}
\newblock \emph{Journal of Personality and Social Psychology}, 39(6):1161--1178.

\bibitem[{Soriano and Ogarkova(2025)}]{soriano-frustration}
Cristina Soriano and Anna Ogarkova. 2025.
\newblock \href {https://doi.org/10.1017/langcog.2024.50} {The meaning of ‘frustration’ across languages}.
\newblock \emph{Language and Cognition}, 17:e16.

\bibitem[{Stadthagen-Gonzalez et~al.(2016)Stadthagen-Gonzalez, Imbault, Pérez~Sánchez, and Brysbaert}]{gonzalez-vad}
Hans Stadthagen-Gonzalez, Constance Imbault, Miguel~A. Pérez~Sánchez, and Marc Brysbaert. 2016.
\newblock \href {https://doi.org/10.3758/s13428-015-0700-2} {Norms of valence and arousal for 14, 031 spanish words}.
\newblock \emph{Behavior Research Methods}, 49(1):111–123.

\bibitem[{Sun et~al.(2026)Sun, Yan, Lu, Lee, Zhang, and Shao}]{sun-subspace}
Lihao Sun, Lewen Yan, Xiaoya Lu, Andrew Lee, Jie Zhang, and Jing Shao. 2026.
\newblock \href {https://arxiv.org/abs/2604.03147} {Valence-arousal subspace in llms: Circular emotion geometry and multi-behavioral control}.
\newblock \emph{Preprint}, arXiv:2604.03147.

\bibitem[{Team et~al.(2025)Team, Kamath, Ferret, Pathak, Vieillard, Merhej, Perrin, Matejovicova, Ramé, Rivière, Rouillard, Mesnard, Cideron, bastien Grill, Ramos, Yvinec, Casbon, Pot, Penchev, Liu, Visin, Kenealy, Beyer, Zhai, Tsitsulin, Busa-Fekete, Feng, Sachdeva, Coleman, Gao, Mustafa, Barr, Parisotto, Tian, Eyal, Cherry, Peter, Sinopalnikov, Bhupatiraju, Agarwal, Kazemi, Malkin, Kumar, Vilar, Brusilovsky, Luo, Steiner, Friesen, Sharma, Sharma, Gilady, Goedeckemeyer, Saade, Feng, Kolesnikov, Bendebury, Abdagic, Vadi, György, Pinto, Das, Bapna, Miech, Yang, Paterson, Shenoy, Chakrabarti, Piot, Wu, Shahriari, Petrini, Chen, Lan, Choquette-Choo, Carey, Brick, Deutsch, Eisenbud, Cattle, Cheng, Paparas, Sreepathihalli, Reid, Tran, Zelle, Noland, Huizenga, Kharitonov, Liu, Amirkhanyan, Cameron, Hashemi, Klimczak-Plucińska, Singh, Mehta, Lehri, Hazimeh, Ballantyne, Szpektor, Nardini, Pouget-Abadie, Chan, Stanton, Wieting, Lai, Orbay, Fernandez, Newlan, yeong Ji, Singh, Black, Yu, Hui, Vodrahalli, Greff, Qiu,
  Valentine, Coelho, Ritter, Hoffman, Watson, Chaturvedi, Moynihan, Ma, Babar, Noy, Byrd, Roy, Momchev, Chauhan, Sachdeva, Bunyan, Botarda, Caron, Rubenstein, Culliton, Schmid, Sessa, Xu, Stanczyk, Tafti, Shivanna, Wu, Pan, Rokni, Willoughby, Vallu, Mullins, Jerome, Smoot, Girgin, Iqbal, Reddy, Sheth, Põder, Bhatnagar, Panyam, Eiger, Zhang, Liu, Yacovone, Liechty, Kalra, Evci, Misra, Roseberry, Feinberg, Kolesnikov, Han, Kwon, Chen, Chow, Zhu, Wei, Egyed, Cotruta, Giang, Kirk, Rao, Black, Babar, Lo, Moreira, Martins, Sanseviero, Gonzalez, Gleicher, Warkentin, Mirrokni, Senter, Collins, Barral, Ghahramani, Hadsell, Matias, Sculley, Petrov, Fiedel, Shazeer, Vinyals, Dean, Hassabis, Kavukcuoglu, Farabet, Buchatskaya, Alayrac, Anil, Dmitry, Lepikhin, Borgeaud, Bachem, Joulin, Andreev, Hardin, Dadashi, and Hussenot}]{gemmateam2025gemma3technicalreport}
Gemma Team, Aishwarya Kamath, Johan Ferret, Shreya Pathak, Nino Vieillard, Ramona Merhej, Sarah Perrin, Tatiana Matejovicova, Alexandre Ramé, Morgane Rivière, Louis Rouillard, Thomas Mesnard, Geoffrey Cideron, Jean bastien Grill, Sabela Ramos, Edouard Yvinec, Michelle Casbon, Etienne Pot, Ivo Penchev, Gaël Liu, Francesco Visin, Kathleen Kenealy, Lucas Beyer, Xiaohai Zhai, Anton Tsitsulin, Robert Busa-Fekete, Alex Feng, Noveen Sachdeva, Benjamin Coleman, Yi~Gao, Basil Mustafa, Iain Barr, Emilio Parisotto, David Tian, Matan Eyal, Colin Cherry, Jan-Thorsten Peter, Danila Sinopalnikov, Surya Bhupatiraju, Rishabh Agarwal, Mehran Kazemi, Dan Malkin, Ravin Kumar, David Vilar, Idan Brusilovsky, Jiaming Luo, Andreas Steiner, Abe Friesen, Abhanshu Sharma, Abheesht Sharma, Adi~Mayrav Gilady, Adrian Goedeckemeyer, Alaa Saade, Alex Feng, Alexander Kolesnikov, Alexei Bendebury, Alvin Abdagic, Amit Vadi, András György, André~Susano Pinto, Anil Das, Ankur Bapna, Antoine Miech, Antoine Yang, Antonia Paterson, Ashish
  Shenoy, Ayan Chakrabarti, Bilal Piot, Bo~Wu, Bobak Shahriari, Bryce Petrini, Charlie Chen, Charline~Le Lan, Christopher~A. Choquette-Choo, CJ~Carey, Cormac Brick, Daniel Deutsch, Danielle Eisenbud, Dee Cattle, Derek Cheng, Dimitris Paparas, Divyashree~Shivakumar Sreepathihalli, Doug Reid, Dustin Tran, Dustin Zelle, Eric Noland, Erwin Huizenga, Eugene Kharitonov, Frederick Liu, Gagik Amirkhanyan, Glenn Cameron, Hadi Hashemi, Hanna Klimczak-Plucińska, Harman Singh, Harsh Mehta, Harshal~Tushar Lehri, Hussein Hazimeh, Ian Ballantyne, Idan Szpektor, Ivan Nardini, Jean Pouget-Abadie, Jetha Chan, Joe Stanton, John Wieting, Jonathan Lai, Jordi Orbay, Joseph Fernandez, Josh Newlan, Ju~yeong Ji, Jyotinder Singh, Kat Black, Kathy Yu, Kevin Hui, Kiran Vodrahalli, Klaus Greff, Linhai Qiu, Marcella Valentine, Marina Coelho, Marvin Ritter, Matt Hoffman, Matthew Watson, Mayank Chaturvedi, Michael Moynihan, Min Ma, Nabila Babar, Natasha Noy, Nathan Byrd, Nick Roy, Nikola Momchev, Nilay Chauhan, Noveen Sachdeva, Oskar
  Bunyan, Pankil Botarda, Paul Caron, Paul~Kishan Rubenstein, Phil Culliton, Philipp Schmid, Pier~Giuseppe Sessa, Pingmei Xu, Piotr Stanczyk, Pouya Tafti, Rakesh Shivanna, Renjie Wu, Renke Pan, Reza Rokni, Rob Willoughby, Rohith Vallu, Ryan Mullins, Sammy Jerome, Sara Smoot, Sertan Girgin, Shariq Iqbal, Shashir Reddy, Shruti Sheth, Siim Põder, Sijal Bhatnagar, Sindhu~Raghuram Panyam, Sivan Eiger, Susan Zhang, Tianqi Liu, Trevor Yacovone, Tyler Liechty, Uday Kalra, Utku Evci, Vedant Misra, Vincent Roseberry, Vlad Feinberg, Vlad Kolesnikov, Woohyun Han, Woosuk Kwon, Xi~Chen, Yinlam Chow, Yuvein Zhu, Zichuan Wei, Zoltan Egyed, Victor Cotruta, Minh Giang, Phoebe Kirk, Anand Rao, Kat Black, Nabila Babar, Jessica Lo, Erica Moreira, Luiz~Gustavo Martins, Omar Sanseviero, Lucas Gonzalez, Zach Gleicher, Tris Warkentin, Vahab Mirrokni, Evan Senter, Eli Collins, Joelle Barral, Zoubin Ghahramani, Raia Hadsell, Yossi Matias, D.~Sculley, Slav Petrov, Noah Fiedel, Noam Shazeer, Oriol Vinyals, Jeff Dean, Demis Hassabis,
  Koray Kavukcuoglu, Clement Farabet, Elena Buchatskaya, Jean-Baptiste Alayrac, Rohan Anil, Dmitry, Lepikhin, Sebastian Borgeaud, Olivier Bachem, Armand Joulin, Alek Andreev, Cassidy Hardin, Robert Dadashi, and Léonard Hussenot. 2025.
\newblock \href {https://arxiv.org/abs/2503.19786} {Gemma 3 technical report}.
\newblock \emph{Preprint}, arXiv:2503.19786.

\bibitem[{Timkey and van Schijndel(2021)}]{timkey-van-schijndel-2021-bark}
William Timkey and Marten van Schijndel. 2021.
\newblock \href {https://doi.org/10.18653/v1/2021.emnlp-main.372} {All bark and no bite: Rogue dimensions in transformer language models obscure representational quality}.
\newblock In \emph{Proceedings of the 2021 Conference on Empirical Methods in Natural Language Processing}, pages 4527--4546, Online and Punta Cana, Dominican Republic. Association for Computational Linguistics.

\bibitem[{van Osch et~al.(2013)}]{vanosch2013pride}
Yvette M.~J. van Osch et~al. 2013.
\newblock \href {https://doi.org/10.1093/acprof:oso/9780199592746.003.0026} {The meaning of pride across cultures}.
\newblock In Johnny J.~R. Fontaine, Klaus~R. Scherer, and Cristina Soriano, editors, \emph{Components of Emotional Meaning: A Sourcebook}. Oxford University Press, Oxford.

\bibitem[{Voita and Titov(2020)}]{voita-titov-2020-information}
Elena Voita and Ivan Titov. 2020.
\newblock \href {https://doi.org/10.18653/v1/2020.emnlp-main.14} {Information-theoretic probing with minimum description length}.
\newblock In \emph{Proceedings of the 2020 Conference on Empirical Methods in Natural Language Processing (EMNLP)}, pages 183--196, Online. Association for Computational Linguistics.

\bibitem[{Wierzbicka(1999)}]{wierzbicka-emotions}
Anna Wierzbicka. 1999.
\newblock \href {https://doi.org/10.1017/cbo9780511521256} {\emph{Emotions across Languages and Cultures: Diversity and Universals}}.
\newblock Cambridge University Press.

\bibitem[{Wu et~al.(2025)Wu, Yu, Yogatama, Lu, and Kim}]{wu2025semantic}
Zhaofeng Wu, Xinyan Yu, Dani Yogatama, Jiasen Lu, and Yoon Kim. 2025.
\newblock \href {https://proceedings.iclr.cc/paper_files/paper/2025/file/84fb08ae30ab266953d1ff72eb0611b1-Paper-Conference.pdf} {The semantic hub hypothesis: Language models share semantic representations across languages and modalities}.
\newblock In \emph{International Conference on Learning Representations (ICLR)}, volume 2025, pages 53705--53723.

\bibitem[{Xu and Zhang(2026)}]{xu-zhang-2026-language}
Shaoyang Xu and Wenxuan Zhang. 2026.
\newblock \href {https://doi.org/10.18653/v1/2026.acl-long.628} {Language of thought shapes output diversity in large language models}.
\newblock In \emph{Proceedings of the 64th Annual Meeting of the {A}ssociation for {C}omputational {L}inguistics (Volume 1: Long Papers)}, pages 13802--13816, San Diego, California, United States. Association for Computational Linguistics.

\bibitem[{Yang et~al.(2025)Yang, Li, Yang, Zhang, Hui, Zheng, Yu, Gao, Huang, Lv, Zheng, Liu, Zhou, Huang, Hu, Ge, Wei, Lin, Tang, Yang, Tu, Zhang, Yang, Yang, Zhou, Zhou, Lin, Dang, Bao, Yang, Yu, Deng, Li, Xue, Li, Zhang, Wang, Zhu, Men, Gao, Liu, Luo, Li, Tang, Yin, Ren, Wang, Zhang, Ren, Fan, Su, Zhang, Zhang, Wan, Liu, Wang, Cui, Zhang, Zhou, and Qiu}]{qwen3}
An~Yang, Anfeng Li, Baosong Yang, Beichen Zhang, Binyuan Hui, Bo~Zheng, Bowen Yu, Chang Gao, Chengen Huang, Chenxu Lv, Chujie Zheng, Dayiheng Liu, Fan Zhou, Fei Huang, Feng Hu, Hao Ge, Haoran Wei, Huan Lin, Jialong Tang, Jian Yang, Jianhong Tu, Jianwei Zhang, Jianxin Yang, Jiaxi Yang, Jing Zhou, Jingren Zhou, Junyang Lin, Kai Dang, Keqin Bao, Kexin Yang, Le~Yu, Lianghao Deng, Mei Li, Mingfeng Xue, Mingze Li, Pei Zhang, Peng Wang, Qin Zhu, Rui Men, Ruize Gao, Shixuan Liu, Shuang Luo, Tianhao Li, Tianyi Tang, Wenbiao Yin, Xingzhang Ren, Xinyu Wang, Xinyu Zhang, Xuancheng Ren, Yang Fan, Yang Su, Yichang Zhang, Yinger Zhang, Yu~Wan, Yuqiong Liu, Zekun Wang, Zeyu Cui, Zhenru Zhang, Zhipeng Zhou, and Zihan Qiu. 2025.
\newblock \href {https://arxiv.org/abs/2505.09388} {Qwen3 technical report}.
\newblock \emph{Preprint}, arXiv:2505.09388.

\bibitem[{Yu et~al.(2016{\natexlab{a}})Yu, Lee, Hao, Wang, He, Hu, Lai, and Zhang}]{yu2016building}
Liang-Chih Yu, Lung-Hao Lee, Shuai Hao, Jin Wang, Yunchao He, Jun Hu, K.~Robert Lai, and Xuejie Zhang. 2016{\natexlab{a}}.
\newblock Building chinese affective resources in valence-arousal dimensions.
\newblock In \emph{Proceedings of the 2016 Conference of the North American Chapter of the Association for Computational Linguistics: Human Language Technologies}, pages 540--545.

\bibitem[{Yu et~al.(2016{\natexlab{b}})Yu, Lee, Hao, Wang, He, Hu, Lai, and Zhang}]{yu-cvaw}
Liang-Chih Yu, Lung-Hao Lee, Shuai Hao, Jin Wang, Yunchao He, Jun Hu, K.~Robert Lai, and Xuejie Zhang. 2016{\natexlab{b}}.
\newblock \href {https://doi.org/10.18653/v1/N16-1066} {Building {C}hinese affective resources in valence-arousal dimensions}.
\newblock In \emph{Proceedings of the 2016 Conference of the North {A}merican Chapter of the Association for Computational Linguistics: Human Language Technologies}, pages 540--545, San Diego, California. Association for Computational Linguistics.

\end{thebibliography}

\appendix
    \section{Colexification Network Replication}

\label{app:jackson-repl}

    \begin{figure}[h]

        \centering
        \includegraphics[width=0.95\linewidth, trim=.25cm 1cm .25cm 0cm]{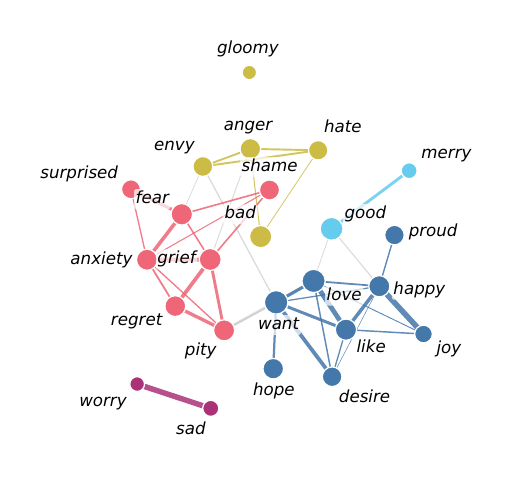}
        \caption{Replication of \citet{jackson-semantics}'s universal colexification network.}
        \label{fig:jackson-colex-network}
    \end{figure}

\begin{table}[h]
    \centering
    \small
    \begin{tabular}{llll}
        \toprule
        \multicolumn{4}{c}{Jackson et al.\ emotion concepts} \\
        \midrule
        anger    & envy   & hate & pity      \\
        anxiety  & fear   & hope & proud     \\
        bad      & gloomy & joy  & regret    \\
        desire   & good   & like & sad       \\
        love     & grief  & merry & shame    \\
        happy    & want   & worry & surprised \\
        \bottomrule
    \end{tabular}
    \caption{\citet{jackson-semantics}'s 24 emotion concepts used to build colexification networks.}
    \label{tab:jackson-concepts}
\end{table}

\section{Psychometric Lexicon Details}
\label{app:lexicon-dets}

\noindent\begin{minipage}{\linewidth}
    \makeatletter\def\@captype{table}\makeatother
    \centering
    \resizebox{\linewidth}{!}{
        \begin{tabular}{cccccccc}
             \toprule
             \multirow{2}{*}{Lang} & \multirow{2}{*}{Dict} & \multirow{2}{*}{\# Terms} & \multirow{2}{*}{\% Cov} & \multicolumn{2}{c}{Val} & \multicolumn{2}{c}{Arou} \\
             \cmidrule(lr){5-6}\cmidrule(lr){7-8}
             & & & & $\mu$ & $\sigma$ & $\mu$ & $\sigma$ \\
             \midrule
             \texttt{en} & \makecell{NRC-VAD v2.1 \\\small\citep{mohammad-vad-v2}} & 44,728 & 98.5 & .40 & .35 & .58 & .22 \\
             \texttt{es} & \makecell{Stadthagen-Gonz\'alez \\\small\citep{gonzalez-vad}} & 14,031 & 80.0 & .46 & .28 & .61 & .17 \\
             \texttt{fr} & \makecell{MEmoLon \\\small\citep{buechel-memolon}} & 2,001,517 & 100.0 & .49 & .06 & .16 & .03 \\
             \texttt{zh} & \makecell{CVAW \\\small\citep{yu-cvaw}} & 5,512 & 89.0 & .40 & .21 & .57 & .18 \\
             \bottomrule
        \end{tabular}
    }
    \caption{VA lexicons used in the confirmatory analysis with summary statistics percent of affective states covered by the lexicon (\% Cov) and the distributions of corresponding valence and arousal scores, rescaled from 0 to 1. Due to data availability and coverage of our data, the lexicon for French is not strictly native-speaker annotated. As a result, it is also much larger, including annotations for 2,001,517 French words.}
    \label{tab:lexicon-stats}
\end{minipage}
\vspace{\baselineskip}

\clearpage
\section{Monolingual networks}
\label{app:monolingual-networks}

\begin{table}[h]
    \centering
    \begin{CJK}{UTF8}{gbsn}
    \resizebox{.95\linewidth}{!}{
    \begin{tabular}{cp{.75\linewidth}}
    \toprule
    \textbf{Lang.} & \textbf{Representative words} \\
    \midrule
    
    \multicolumn{2}{c}{\textbf{Positive affect}} \\
    \midrule
    
    \texttt{en} &
    happy, energized, good, wonderful, empowered \\
    
    \texttt{sp} &
    complacido\gl{pleased}, contento/a\gl{content}, satisfecho/a\gl{satisfied},
    ilusionado\gl{hopeful}, entusiasmado/a\gl{enthusiastic} \\
    
    \texttt{fr} &
    libéré(e)\gl{freed}, apaisé(e)\gl{soothed}, détendue\gl{relaxed},
    délivré(e)\gl{relieved}, joyeuse\gl{joyful} \\
    
    \texttt{ch} &
    开心\gl{happy}, 喜悦\gl{joy}, 愉悦\gl{pleased},
    兴奋\gl{excited}, 欣喜\gl{delighted} \\
    
    \midrule
    \multicolumn{2}{c}{\textbf{Negative affect}} \\
    \midrule
    
    \texttt{en} &
    upset, distressed, panicked, irritated, devastated \\
    
    \texttt{sp} &
    desanimado/a\gl{discouraged}, agotado/a\gl{exhausted}, agobiado\gl{overwhelmed},
    destrozado/a\gl{shattered}, desesperado/a\gl{desperate} \\
    
    \texttt{fr} &
    désemparé(e)\gl{helpless}, dégoûté(e)\gl{disgusted}, énervé\gl{irritated},
    inquiète\gl{worried}, effrayé(e)\gl{frightened} \\
    
    \texttt{ch} &
    悲伤\gl{sorrowful}, 难过\gl{sad}, 伤心\gl{heartbroken},
    沮丧\gl{dejected}, 伤感\gl{wistful} \\
    
    \bottomrule
    \end{tabular}
    }
    \end{CJK}
    \caption{Louvain community modularity on the top 20\% of edges Qwen3-8B-Base, layer 18, detected separately per language. Masculine and feminine forms are merged.}
    \label{tab:monolingual_communities}
\end{table}

\section{Pride case study}
 We create monolingual proximity networks in the same manner as for `frustration,' and evaluate to what extent `proud' and its translations in French, Spanish, and Mandarin demonstrate cross-lingual variation in semantically related emotion terms. In Table \ref{tab:pride_neighbors}, for Qwen3 and Gemma3, the semantically closest emotion word representations to English `pride' are all positive, for example, `glad' and `happy.' Ministral3 has only one negative related word: `ashamed.' On the other hand, all of the Mandarin closest semantic neighbors include negative concepts, some examples including `ashamed' and `guilty.'
\label{app:pride_case_study}

\begin{table}[tp]
\centering
\begin{CJK}{UTF8}{gbsn}
\setlength{\tabcolsep}{5pt}
\renewcommand{\arraystretch}{0.92}

\begin{subtable}[t]{\columnwidth}
\centering
\scriptsize
\begin{tabular}{@{}ll@{}}
\toprule
\textbf{proud} & \textbf{自豪} \\
\midrule
\makecell[l]{glad \\ {\tiny\textcolor{gray}{(0.562)}}} &
\makecell[l]{骄傲 \\ {\tiny\textcolor{gray}{(proud/arrogant; 0.737)}}} \\
\makecell[l]{accomplished \\ {\tiny\textcolor{gray}{(0.549)}}} &
\makecell[l]{欣慰 \\ {\tiny\textcolor{gray}{(gratified; 0.548)}}} \\
\makecell[l]{happy \\ {\tiny\textcolor{gray}{(0.528)}}} &
\makecell[l]{荣幸 \\ {\tiny\textcolor{gray}{(honored; 0.493)}}} \\
\makecell[l]{grateful \\ {\tiny\textcolor{gray}{(0.499)}}} &
\makecell[l]{高兴 \\ {\tiny\textcolor{gray}{(glad; 0.459)}}} \\
\makecell[l]{excited \\ {\tiny\textcolor{gray}{(0.499)}}} &
\makecell[l]{满足 \\ {\tiny\textcolor{gray}{(satisfied; 0.453)}}} \\
\bottomrule
\end{tabular}
\caption{Qwen3-8B-Base, Layer 18}
\end{subtable}

\vspace{6pt}

\begin{subtable}[t]{\columnwidth}
\centering
\scriptsize
\begin{tabular}{@{}ll@{}}
\toprule
\textbf{proud} & \textbf{自豪} \\
\midrule
\makecell[l]{glad \\ {\tiny\textcolor{gray}{(0.648)}}} &
\makecell[l]{骄傲 \\ {\tiny\textcolor{gray}{(proud/arrogant; 0.659)}}} \\
\makecell[l]{happy \\ {\tiny\textcolor{gray}{(0.613)}}} &
\makecell[l]{欣慰 \\ {\tiny\textcolor{gray}{(gratified; 0.533)}}} \\
\makecell[l]{honored \\ {\tiny\textcolor{gray}{(0.601)}}} &
\makecell[l]{荣幸 \\ {\tiny\textcolor{gray}{(honored; 0.491)}}} \\
\makecell[l]{excited \\ {\tiny\textcolor{gray}{(0.596)}}} &
\makecell[l]{自信 \\ {\tiny\textcolor{gray}{(confident; 0.478)}}} \\
\makecell[l]{ashamed \\ {\tiny\textcolor{gray}{(0.588)}}} &
\makecell[l]{羞愧 \\ {\tiny\textcolor{gray}{(ashamed; 0.477)}}} \\
\bottomrule
\end{tabular}
\caption{Ministral-3-8B-Base, Layer 17}
\end{subtable}

\vspace{6pt}

\begin{subtable}[t]{\columnwidth}
\centering
\scriptsize
\begin{tabular}{@{}ll@{}}
\toprule
\textbf{proud} & \textbf{自豪} \\
\midrule
\makecell[l]{glad \\ {\tiny\textcolor{gray}{(0.723)}}} &
\makecell[l]{骄傲 \\ {\tiny\textcolor{gray}{(proud/arrogant; 0.723)}}} \\
\makecell[l]{grateful \\ {\tiny\textcolor{gray}{(0.687)}}} &
\makecell[l]{欣慰 \\ {\tiny\textcolor{gray}{(gratified; 0.691)}}} \\
\makecell[l]{happy \\ {\tiny\textcolor{gray}{(0.687)}}} &
\makecell[l]{荣幸 \\ {\tiny\textcolor{gray}{(honored; 0.612)}}} \\
\makecell[l]{honored \\ {\tiny\textcolor{gray}{(0.659)}}} &
\makecell[l]{羞愧 \\ {\tiny\textcolor{gray}{(ashamed; 0.589)}}} \\
\makecell[l]{excited \\ {\tiny\textcolor{gray}{(0.657)}}} &
\makecell[l]{内疚 \\ {\tiny\textcolor{gray}{(guilty; 0.575)}}} \\
\bottomrule
\end{tabular}
\caption{Gemma-3-12B-pt, Layer 24}
\end{subtable}

\caption{Five nearest neighbors for pride in English and Mandarin
across models, using standardized cosine similarity.
Parentheses give English glosses and cosine similarity.}
\label{tab:pride_neighbors}
\end{CJK}
\end{table}

\label{app:multilingual-networks}

\begin{table*}[!t]
\section{Multilingual networks}

\centering
\begin{CJK}{UTF8}{gbsn}
\resizebox{\textwidth}{!}{
\begin{tabular}{lp{.20\linewidth}p{.24\linewidth}p{.24\linewidth}p{.20\linewidth}}
\toprule
\textbf{Community} & \texttt{en} & \texttt{sp} & \texttt{fr} & \texttt{ch} \\
\midrule

Positive &
happy, grateful, proud &
feliz\gl{happy}, agradecido\gl{grateful}, orgulloso\gl{proud} &
heureux\gl{happy}, reconnaissant\gl{grateful}, fier\gl{proud} &
开心\gl{happy}, 感恩\gl{grateful}, 自豪\gl{proud} \\

Depletion / despair &
depressed, exhausted, hopeless &
deprimido\gl{depressed}, agotado\gl{exhausted}, desesperado\gl{desperate} &
déprimé\gl{depressed}, fatigué\gl{tired}, désespéré\gl{desperate} &
抑郁\gl{depressed}, 疲惫\gl{exhausted}, 绝望\gl{despair} \\

Anger / sadness &
angry, frustrated, sad &
enojado\gl{angry}, frustrado\gl{frustrated}, triste\gl{sad} &
en colère\gl{angry}, frustré\gl{frustrated}, triste\gl{sad} &
生气\gl{angry}, 沮丧\gl{dejected}, 伤心\gl{sad} \\

Fear / anxiety &
afraid, anxious, worried &
asustado\gl{scared}, ansioso\gl{anxious}, preocupado\gl{worried} &
effrayé\gl{frightened}, angoissé\gl{anguished}, inquiet\gl{worried} &
害怕\gl{afraid}, 焦虑\gl{anxious}, 担心\gl{worried} \\

Shame / guilt &
ashamed, guilty, embarrassed &
avergonzado\gl{ashamed}, culpable\gl{guilty}, arrepentido\gl{regretful} &
coupable\gl{guilty}, gêné\gl{embarrassed}, désolé\gl{sorry} &
羞愧\gl{ashamed}, 内疚\gl{guilty}, 尴尬\gl{embarrassed} \\

Treatment by others &
betrayed, rejected, loved &
traicionado\gl{betrayed}, engañada\gl{deceived}, querido\gl{loved} &
rejeté\gl{rejected}, abandonné\gl{abandoned}, aimé\gl{loved} &
冷漠\gl{indifferent}, 有爱\gl{loving} \\

\bottomrule
\end{tabular}
}
\end{CJK}
\caption{Multilingual communities for Qwen3-8B-Base, layer 18 resulting from Louvain community detection on the top 20\% of edges after residualizing out pairwise similarity.}
\label{tab:joint_communities}
\end{table*}

\noindent\begin{minipage}{\columnwidth}
    \makeatletter\def\@captype{figure}\makeatother
    \centering
    \includegraphics[width=\linewidth]{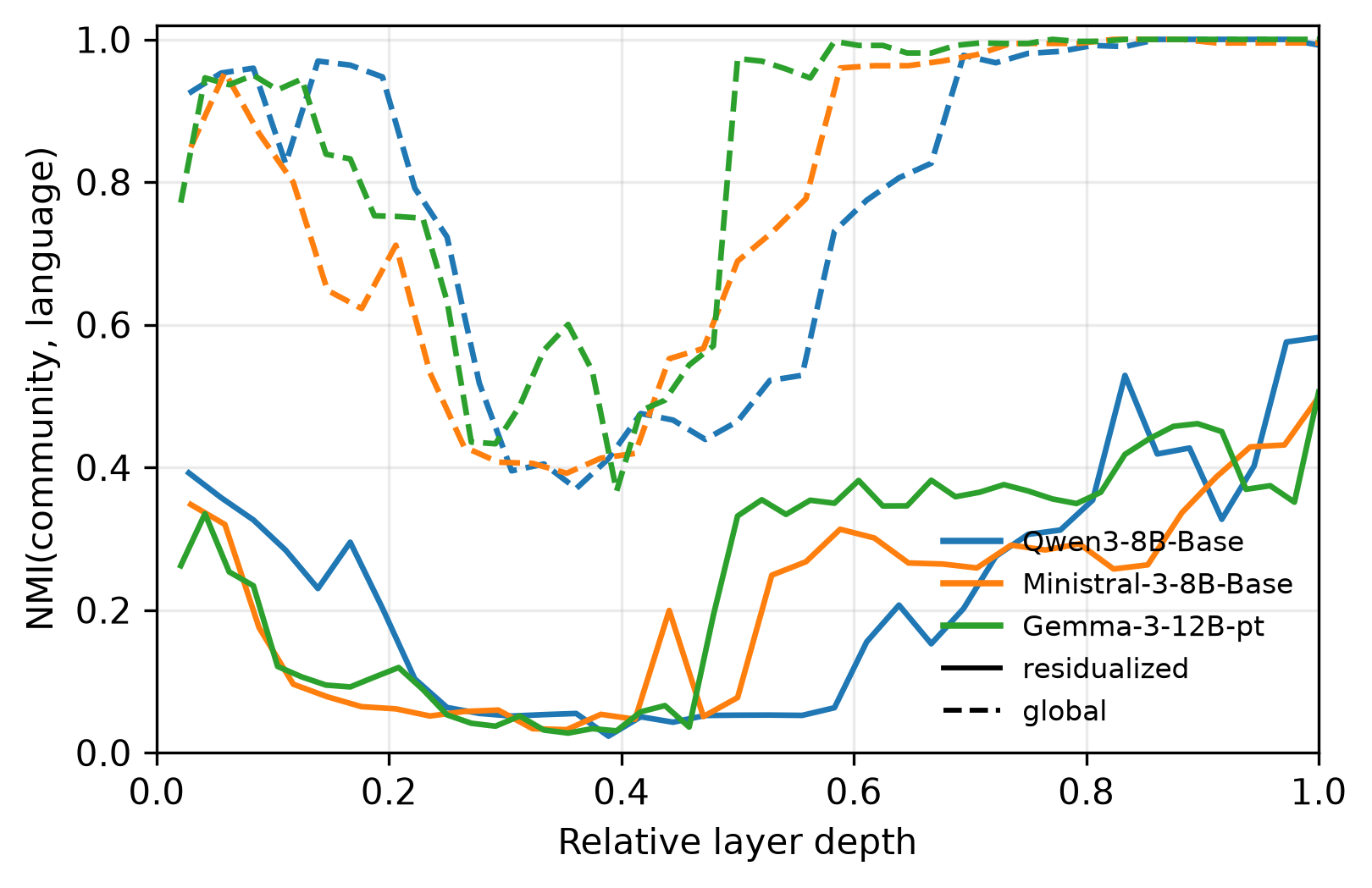}
    \caption{NMI community separation for multilingual network before and after partialling out language similarity.}
    \label{fig:nmi}
\end{minipage}
\vspace{\baselineskip}

\noindent\begin{minipage}{\columnwidth}
    \makeatletter\def\@captype{figure}\makeatother
    \centering
    \includegraphics[width=\linewidth]{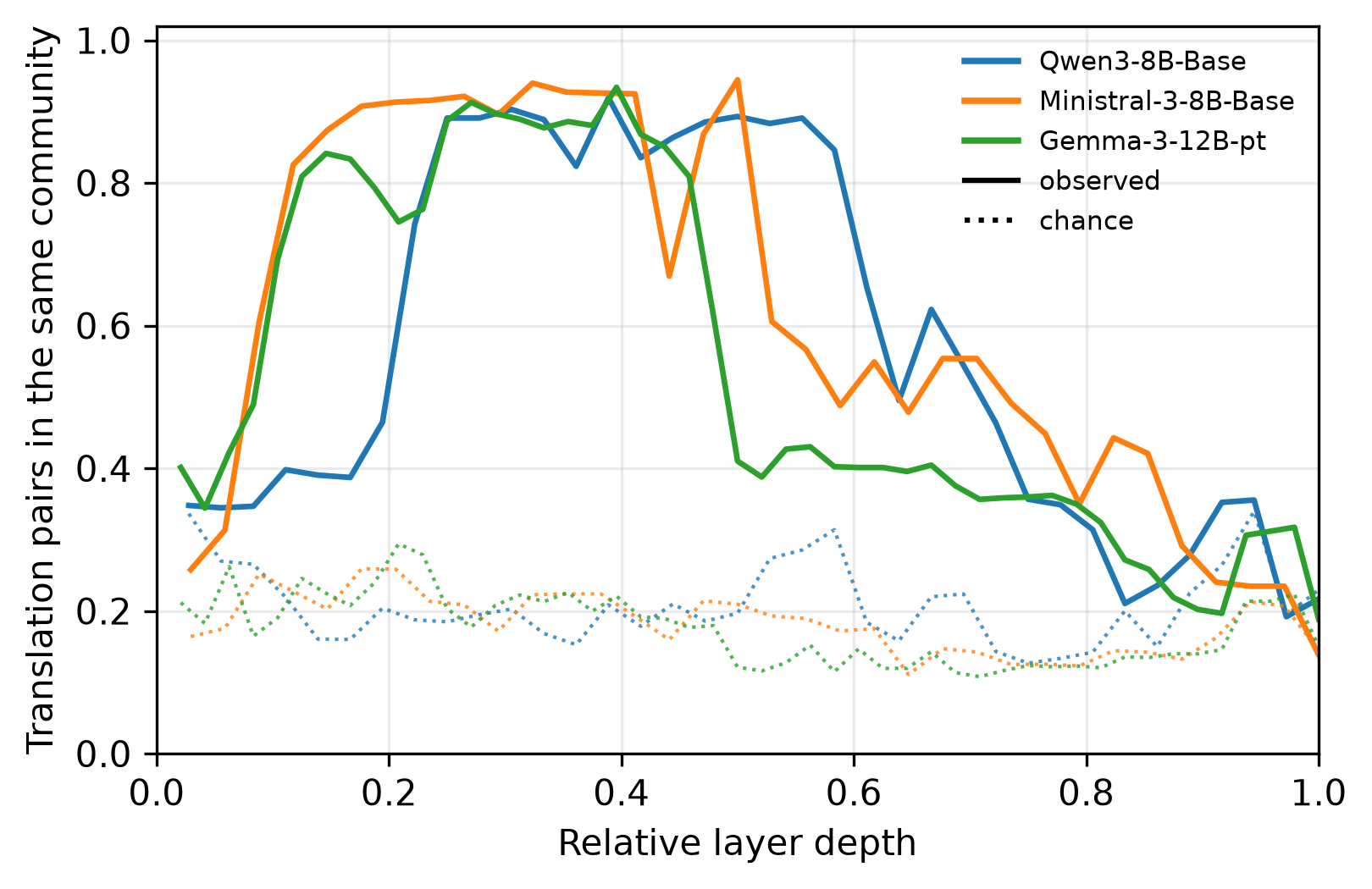}
    \caption{Proportion of translation pairs, as determined by each model's generated translations, falling in the same Louvain-detected community in each layer.}
    \label{fig:translation_pairs}
\end{minipage}
\vspace{\baselineskip}

\clearpage   %

\section{MDL Probing Details}

\label{app:mdl-probing}

We perform MDL probing by considering continuous ridge-regression probes, and so assume a Gaussian error. Given contextual representations, we find the number of average bits per example required to predict labels, corresponding to the `amount of effort' required to extract representational information. In this case, a larger number of bits corresponds to a larger amount of effort, and increased difficulty of predicting labels. Specifically, we use \textit{Online Coding}, which includes the model indirectly through assessment of the minimum number of bits to transmit valence and arousal labels alone \cite{voita-titov-2020-information}.

\end{document}